\documentclass[10pt,twocolumn,letterpaper]{article}

\usepackage{cvpr}
\usepackage{balance}
\usepackage{amsmath}
\usepackage{scalerel}
\usepackage{bm,multirow}
\usepackage{balance}
\usepackage{booktabs}
\usepackage{multirow}
\usepackage{siunitx}
\usepackage{xcolor}
\usepackage{graphicx}
\usepackage{relsize}
\usepackage[table]{xcolor}
\usepackage{float}

\definecolor{cvprblue}{rgb}{0.21,0.49,0.74}
\definecolor{linkblue}{RGB}{55,126,184}
\usepackage[pagebackref,breaklinks,colorlinks,allcolors=cvprblue]{hyperref}

\title{InverFill: One-Step Inversion for Enhanced Few-Step Diffusion Inpainting}

\author{
Duc Vu$^{1\boldsymbol{\star}}$ \quad Kien Nguyen$^{1\boldsymbol{\star}}$ \quad Ngan Nguyen$^{1\boldsymbol{\star}}$ \quad Trong-Tung Nguyen$^{1\boldsymbol{\star}}$ \\
Phong Nguyen$^{1}$ \quad Khoi Nguyen$^{1}$ \quad Cuong Pham$^{1,2}$ \quad Anh Tran$^{1}$ \\
\small{\textsuperscript{1} Qualcomm AI Research$^{\dagger}$ \quad \textsuperscript{2} Posts \& Telecommunications Inst. of Tech., Vietnam
}\\
\texttt{\scriptsize \{ducvu, kienn, ngannguy, tunnguy, phongnh, khoi, pcuong, anhtra\}@qti.qualcomm.com} \quad 
\texttt{\scriptsize cuongpv@ptit.edu.vn}\\
\small{Project Page: \href{https://qualcomm-ai-research.github.io/InverFill}{\textcolor{cvprblue}{https://qualcomm-ai-research.github.io/InverFill}}}
}

\begin{document}
\makeatletter
\g@addto@macro\@maketitle{\vspace{-13mm}
  \begin{figure}[H]
  \setlength{\linewidth}{\textwidth}
  \setlength{\hsize}{\textwidth}
  \centering
  \includegraphics[width=\textwidth]{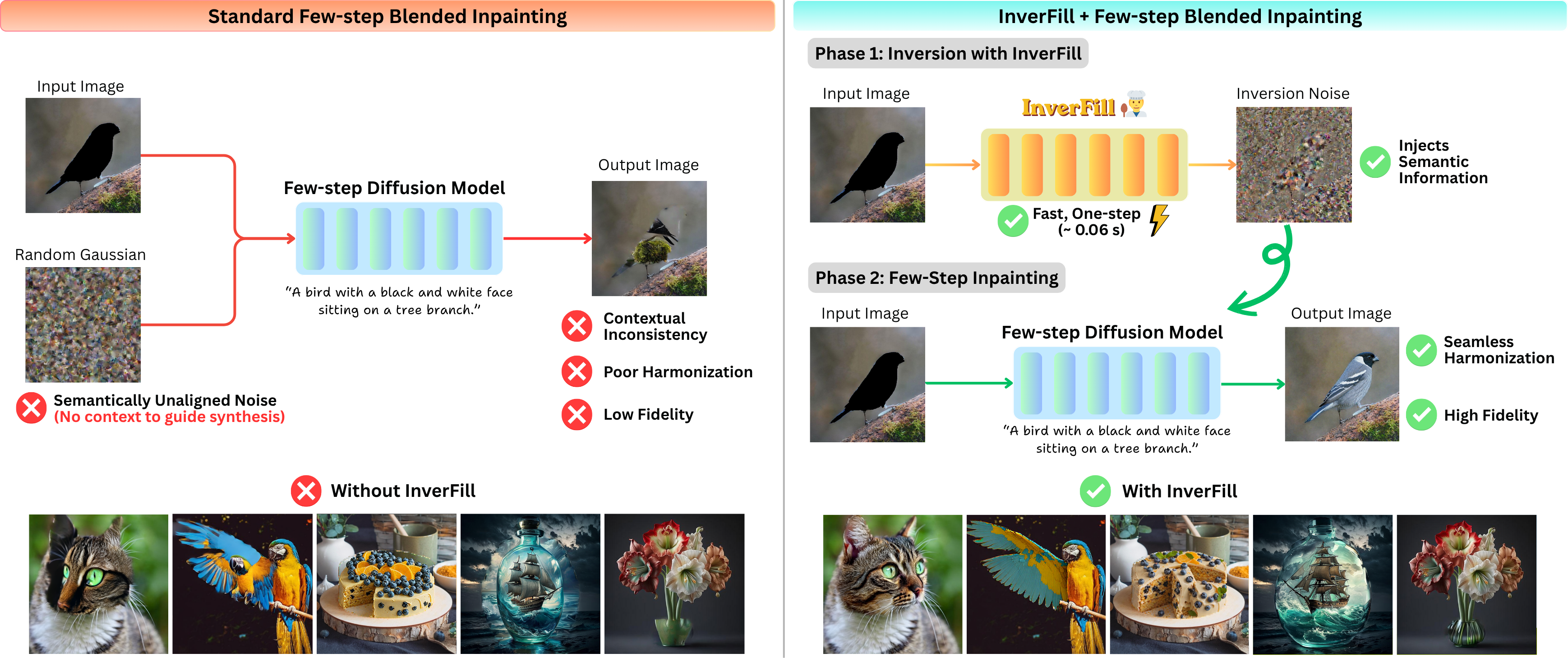}
  \caption{\textbf{InverFill} enhances few-step inpainting by generating semantically aligned inverted noise latents, while adding as little as \textbf{0.06 seconds} of overhead on a single NVIDIA A100 40GB GPU.} 
  \label{fig:teaser}
  \end{figure}
}

\maketitle
\newcommand\blfootnote[1]{%
  \begingroup
  \renewcommand\thefootnote{}\footnote{#1}%
  \addtocounter{footnote}{-1}%
  \endgroup
}

\makeatletter
\def\blfootnote{\gdef\@thefnmark{}\@footnotetext}
\makeatother
    
 \blfootnote{%
  \hspace{-1.7em}$^{\boldsymbol{\star}}$ Equal Contribution \\
  $\dagger$ Qualcomm AI Research is an initiative of Qualcomm Technologies, Inc.%
}

\begin{abstract}
Recent diffusion-based models achieve photorealism in image inpainting but require many sampling steps, limiting practical use. Few-step text-to-image models offer faster generation, but naively applying them to inpainting yields poor harmonization and artifacts between the background and inpainted region. We trace this cause to random Gaussian noise initialization, which under low function evaluations causes semantic misalignment and reduced fidelity. To overcome this, we propose InverFill, a one-step inversion method tailored for inpainting that injects semantic information from the input masked image into the initial noise, enabling high-fidelity few-step inpainting. Instead of training inpainting models, InverFill leverages few-step text-to-image models in a blended sampling pipeline with semantically aligned noise as input, significantly improving vanilla blended sampling and even matching specialized inpainting models at low NFEs. Moreover, InverFill does not require real-image supervision and only adds minimal inference overhead. Extensive experiments show that InverFill consistently boosts baseline few-step models, improving image quality and text coherence without costly retraining or heavy iterative optimization.
\end{abstract}    
\section{Introduction}
\label{sec:intro}
Recent generative models enable photorealistic and detail-rich visual synthesis across many tasks \cite{sdxl,sd,ledit,wang2024sinsr,nguyen2024flexedit}. Among them, text-guided image inpainting has become a key direction, aiming to fill masked regions with content that is semantically aligned with the prompt and visually consistent with the background. Progress in this area is largely driven by large-scale text-to-image diffusion models \cite{sd, flux2024, chenpixart}. To adapt pretrained models for inpainting, recent methods rely on blended sampling or fine-tuning with spatially aware architectures, exploiting strong pretrained priors for seamless results. Early approaches fine-tune the full diffusion U-Net with mask conditioning \cite{sd, nichol2021glide, xie2023smartbrush}, while adapter-based methods like BrushNet \cite{brushnet} add lightweight trainable branches to frozen backbones. Training-free methods \cite{manukyan2023hd, hsiao2024freecond,nullinp} instead use guided sampling or attention manipulation.
Despite their effectiveness, most techniques require many sampling steps, resulting in high latency and limiting real-time deployment. This underscores the need for faster inpainting solutions.

While many few-step text-to-image diffusion models exist \cite{add,ladd,luo2023latent,sanasprint}, adapting them for image inpainting is nontrivial. A natural solution is blended sampling \cite{bld}, where predictions are iteratively merged with the unmasked regions. This strategy works well for multi-step diffusion models, where gradual denoising allows the synthesized content to blend smoothly with the preserved context. In the few-step regime, however, each denoising step induces much larger updates, leading to semantic misalignment between the initial noise and the masked content, and ultimately causing poor harmonization with the surrounding background. To the best of our knowledge, TurboFill \cite{turbofill} is the only successful few-step, text-guided specialized inpainting model. It reduces inference steps using a 3-step adversarial scheme that trains an inpainting adapter on top of a distilled few-step text-to-image model \cite{dmd2}. However, this design is complex, requires real-image supervision, and is computationally heavy. Moreover, prior inpainting methods follow the standard diffusion process \cite{ddpm, ddim}, which always starts from pure Gaussian noise. This gives the model no initial clue about the semantics or structure of the unmasked regions, often causing a semantic mismatch between the inpainted content and its surrounding context. Multi-step models can gradually correct this mismatch, but few-step or one-step models have no such allowance, leaving little room to recover from the initial randomness. As a result, inpainting under low NFEs tends to produce blurry, poorly integrated regions and degraded overall fidelity.

To this end, we introduce \textbf{InverFill}, an efficient one-step inversion network that significantly improves performance of few-step inpainting with minimal overhead. As shown in \cref{fig:teaser}, InverFill maps the masked image into an inverted noise latent, replacing random Gaussian initialization with a semantically informed noise for few-step inpainting. Although diffusion inversion has been explored for editing and inpainting~\cite{Mokady_2023_CVPR, miyake2025negative, corneanu2024latentpaint, nullinp}, we are the first to design a one-step inversion customized for inpainting. While SwiftEdit~\cite{Nguyen_2025_CVPR} proposes a one-step inversion framework for image editing, a naive adaptation to inpainting fails for two reasons. First, training on masked images causes substantial leakage from the visible regions into the inverted noise latent. Second, its reconstruction objective does not constrain the inverted latent to follow the required Gaussian distribution. To overcome this, we introduce Re-Blending to prevent information leakage and a Gaussian regularization loss to ensure the inverted noise latent aligns with the expected noise distribution. Our training pipeline is image-free, requiring no curated image–mask–text triplets and no multi-stage procedures. With these designs, InverFill enhances few-step inpainting and enables few-step text-to-image models to perform on par with specialized inpainting systems, without any finetuning while introducing negligible latency. Our contributions are summarized as follows:
\begin{itemize}
    \item We propose \textbf{InverFill}, an efficient one-step inversion network for few-step image inpainting, which generates semantically informed initial noise to improve inpainting quality while introducing minimal overhead. 
    \item We introduce the \textbf{Re-Blending} operation to mitigate information leakage during training while preserving key semantics in the inverted noise latent for inpainting.
    \item We introduce a \textbf{Gaussian regularization} loss to align the inverted noise latent with the expected Gaussian distribution, enhancing stability and quality.
    \item Our method features a \textbf{highly simplified, image-free training} pipeline that eliminates the need for image-mask-text triplets and complex multi-stage training.
    \item We demonstrate that InverFill significantly boosts the performance of existing few-step inpainting models and enables few-step text-to-image models to perform high-quality inpainting without any task-specific fine-tuning.
\end{itemize}

\section{Related Works}
\label{sec:related_works}
\begin{figure*}[t]
  \centering
  \includegraphics[width=0.99\linewidth]{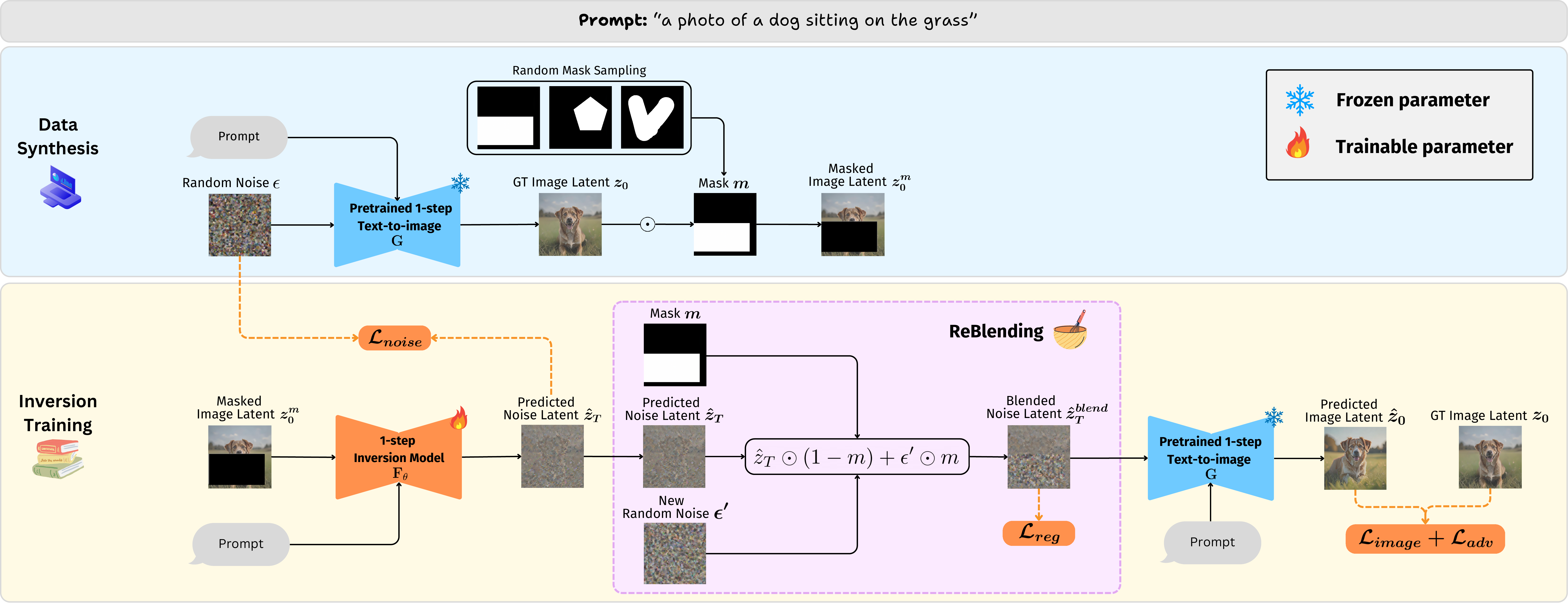}
  \caption{\textbf{Inversion Network Training:} We train an inversion network, $\mathbf{F}_\theta$, to invert a masked image to a noise latent ${\hat{z}_T}$ such that, after blending with random Gaussian noise to form ${\hat{z}}_T^{\text{blend}}$, the latent enables high-fidelity, well-harmonized reconstruction of the original image.}
  \label{fig:inversion_network}
\end{figure*}

\begin{figure}[t]
  \centering
  \includegraphics[width=\linewidth]{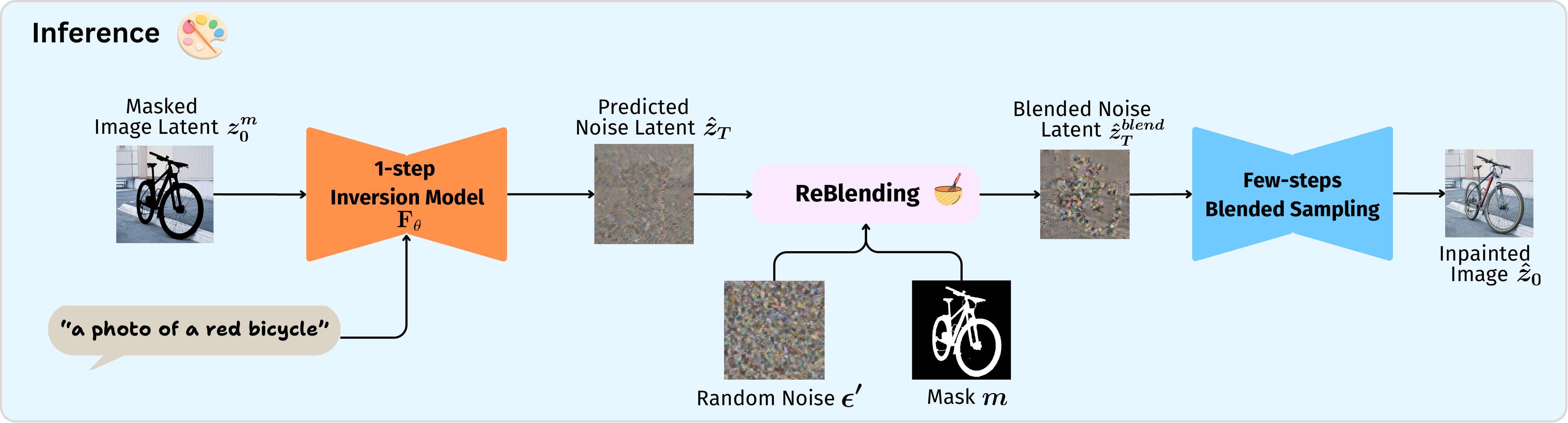}
  \caption{\textbf{Inpainting Pipeline:} The inversion network extracts the latent ${\hat{z}_T}$ from a masked image, which is blended with random noise to form $\hat{z}_T^{\text{blend}}$ and then fed into a few-step inpainting pipeline to generate the final image. \textit{(Zoom in for details)}}
  \label{fig:inpainting_process}
  \vspace{-5mm}
\end{figure}

\subsection{Fast Text-to-image Diffusion Models}
Traditional multi-step diffusion models~\cite{sd,sdxl,chenpixart,ddpm} are known for slow sampling, often requiring dozens to hundreds of neural function evaluations (NFEs) per image. Recent diffusion distillation methods \cite{salimans2022progressive, 10.5555/3618408.3619743, luo2023latent,sb,sbv2,snoopi} significantly accelerate generation by aligning the student model’s prediction trajectory with that of a pre-trained multi-step teacher, enabling few-step (4-8 step) inference. Progressive Distillation \cite{salimans2022progressive} repeatedly distills from the teacher while halving the number of steps at each stage, preserving high sample quality while reducing from thousands of steps. Consistency Model \cite{10.5555/3618408.3619743, luo2023latent,sc} enforces self-consistency in predictions via either distillation-based or distillation-free objectives. ADD \cite{add} and LADD \cite{ladd} employs a combination of adversarial training and score distillation for turning pretrained multi-step diffusion models into few-step diffusion model. SANA-Sprint \cite{sanasprint} accelerates sampling with a training-free transformation into TrigFlow \cite{lu2025simplifying}, followed by few-step training with dense time embeddings, QK-normalization, and max-time weighting.

\subsection{Image Inpainting Approaches}
Image inpainting fills missing regions so they blend naturally with the surrounding context. Early methods \cite{lugmayr2022repaint,corneanu2024latentpaint,preechakul2022diffusion} use GANs or unconditional diffusion models trained on specific datasets \cite{celeba,ffhq,efhq}. For instance, RePaint \cite{lugmayr2022repaint} uses an unconditional DDPM~\cite{ddpm} as a generative prior and blends available pixels into the sampling process. Text-to-image diffusion models provide strong image–text priors for text-guided inpainting, which demands both realistic content completion and semantic alignment with the prompt. BrushNet~\cite{brushnet} fine-tunes both a pretrained text-to-image model and an additional conditional branch for inpainting, and then relies on multi-step sampling to produce coherent results. Meanwhile, Blended Latent Diffusion \cite{bld} guides the multi-step sampling process using a blending operation, gradually aligning the inpainting content with the surrounding background from the source image. Such methods require many NFEs to achieve high-quality results. As fast few-step generative models emerge \cite{luo2023latent, sanasprint, ladd}, reducing the number of sampling steps becomes increasingly necessary, motivating the study of few-step inpainting. A straightforward idea is to apply similar blending strategies on few-step text-to-image models. However, extending inpainting to few-step diffusion models~\cite{luo2023latent, sanasprint, ladd} remains challenging, as blended sampling alone is insufficient to produce coherent results, leading to poor visual quality, as shown in \cref{fig:teaser}. TurboFill~\cite{turbofill} addresses this by training an inpainting adapter on a few-step text-to-image generation model with a complex 3-step adversarial training scheme, which requires extensive real-image supervision. Moreover, TurboFill exclusively explores on UNet–based architectures \cite{sd, sdxl}, leaving its generalization to other models questionable. Hence, few-step inpainting remains under-explored.

\subsection{Diffusion-based Inversion}
While diffusion models generate images by progressively denoising a noisy latent, diffusion inversion methods \cite{ddim, Mokady_2023_CVPR, ju2024pnp, renoise, samuel2025lightningfast} perform the reverse: recovering an inverted latent that faithfully reconstructs the original image when re-denoised. Such inversion is essential for reconstruction, latent exploration, and downstream editing. DDIM Inversion \cite{ddim} introduced a deterministic reverse process by linearizing noise prediction across adjacent steps, an approximation effective for models with many sampling iterations \cite{dhariwal2021diffusion,sd,flux2024,sdxl}. This enables reversed sampling for faithful reconstruction and editing. Null-text Inversion \cite{Mokady_2023_CVPR} refines null-text embeddings via costly iterative optimization, whereas Direct Inversion \cite{ju2024pnp} eliminates this optimization by decoupling reconstruction and editing pathways.

However, the linear approximation used in prior inversion methods breaks down for few-step diffusion models \cite{10.5555/3618408.3619743, luo2023latent, add}, resulting in poor inversion quality. Recent works \cite{renoise,samuel2025lightningfast} therefore develop inversion techniques tailored to few-step models \cite{sdxl, flux2024}. Renoise \cite{renoise} refines noise latents using fixed-point iteration combined with step-wise averaging, while GNRI \cite{samuel2025lightningfast} formulates inversion as a scalar root-finding problem solved with 1-2 Newton–Raphson iterations per step. These methods significantly accelerate and stabilize inversion compared to multi-step approaches. Recently, SwiftEdit \cite{Nguyen_2025_CVPR} pushes this further with a one-step inversion network trained for one-step diffusion models \cite{liu2023instaflow,dmd,dmd2,sb,sbv2}. This network directly maps source images into its noise latent in a single forward pass, enabling fast image reconstruction and editing with minimal overhead. Inspired by this, we incorporate a similar inversion network into our inpainting framework, enhanced with refinements and dedicated training objectives to enable efficient, high-quality few-step inpainting.

\section{Preliminaries}
\subsection{Text-to-Image Diffusion Models.} Text-to-image diffusion models synthesize images by aligning textual inputs with corresponding visual features. State-of-the-art methods primarily use latent diffusion~\cite{sd,sdxl}, where a Variational Auto-Encoder (VAE)~\cite{vae} encoder $\mathcal{E}$ maps an image $I$ to a latent $z$. The denoising process comprises a fixed forward noising step and a learned reverse step. In the forward process, a clean latent $z_0 = \mathcal{E}(I)$ is gradually corrupted into Gaussian noise over $T$ timesteps via a Markov chain $q(z_t | z_{t-1})$ with a variance schedule $\beta_t$:
\begin{equation}
    \label{eq:forward_process}
    q(z_t | z_{t-1}) = \mathcal{N}(z_t; \sqrt{1 - \beta_t} z_{t-1}, \beta_t \mathbf{I}).
\end{equation}
\begin{equation}
    \label{eq:forward_process_reparam}
    z_t = \sqrt{1 - \beta_t} z_{t-1} + \sqrt{\beta_t} \epsilon, \quad \text{where } \epsilon \sim \mathcal{N}(0, \mathbf{I}).
\end{equation}

Given an input noise $z_T$ sampled from \cref{eq:forward_process} and a text prompt $c$, the training objective of the denoising network $\epsilon_\theta$ at timestep $t$ is defined as:
\begin{equation}
\label{eq:simple_objective}
\min_{\theta} \mathbb{E}_{z_0,c, \epsilon \sim \mathcal{U}(1,T), t \sim \mathcal{N}(0, I)} \left\| \epsilon - \epsilon_{\theta}(z_t, t, c) \right\|_2^2
\end{equation}
During inference, $\epsilon_\theta$ iteratively estimates and removes the noise from the noisy image across $T$ timesteps. In practice, large $T$ are required to gradually refine the image, ensuring high-quality generation. In contrast, few-step models apply large, discrete updates at each step, which limits the opportunity for smooth adjustments. Any intermediate modification, such as blending, can easily disrupt the denoising trajectory, leading to artifacts or failed reconstructions.

\subsection{Image Inpainting}
\textbf{Problem Definition.} Given a masked image $I_m \in \mathbb{R}^{H \times W \times C}$ with missing content defined by a binary mask $M \in \{0, 1\}^{H \times W \times C}$, where 0 denotes unmasked regions and 1 denotes masked regions, image inpainting aims to generate $I_{inpaint}$ within the masked region to form a composited image $I = I_m \odot (1 - M) + I_{inpaint} \odot M$ such that the inpainted regions are semantically aligned with a text prompt $c$ and visually consistent with the unmasked context, accurately reflecting what and where to inpaint.

\noindent\textbf{Blended Sampling Strategy.}\label{subsec:blendsample} This inpainting approach, exemplified by Blended Latent Diffusion (BLD) \cite{bld}, gradually blends known information from unmasked regions with generated content in the masked areas. Given a masked image ${I}_m$ and a corresponding binary mask $M$ (resized to $m$ in the latent space), the initial masked latent representation is computed as ${z}_0^m = \mathcal{E}({I}_m)$. During the reverse diffusion process, at each timestep $t$, BLD adds noise to the known regions of the original latent ${z}_0^m$ following \cref{eq:forward_process}, yielding ${z}_t^m$. As described in \cref{eq:blend_ori}, BLD then blends ${z}_t^m$ with the predicted denoised latent $\hat{z}_t$ using the mask $m$. The resulting blended latent serves as the input for the subsequent denoising step at $t-1$, ensuring a seamless transition between the unmasked context and the newly generated content.
\begin{equation}
    z_{t}^{blend} = {z}^m_t \odot {(1-m)} + {\hat{z}}_t \odot m
    \label{eq:blend_ori}
\end{equation}

\begin{figure}[t]
  \centering
  \includegraphics[width=.99\linewidth]{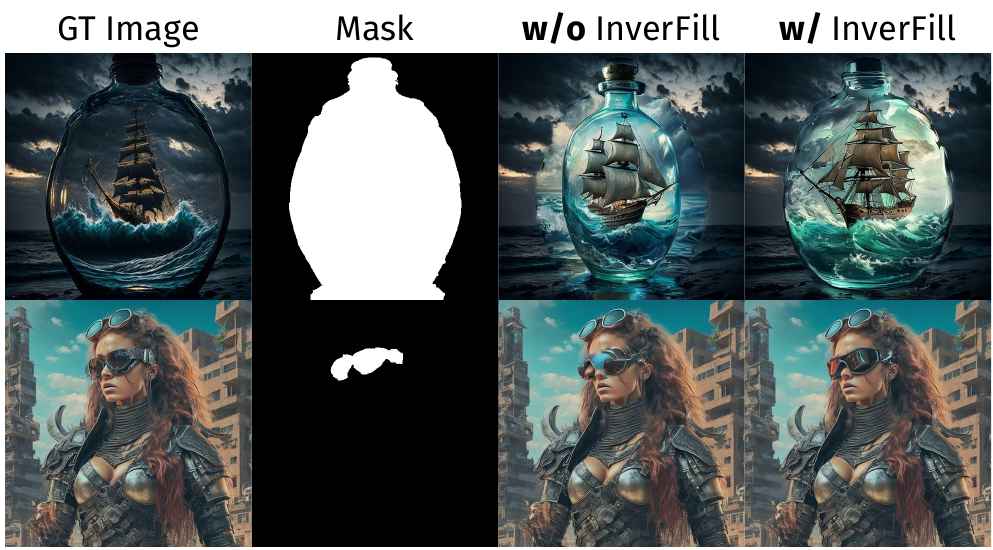}
  \caption{\textbf{Failure of BLD in few-step models} (SDXL-Turbo, 4 steps) is illustrated in \textit{Column 3} and corrected by our method in \textit{Column 4}. \textit{(Zoom in for details)}}
  \label{fig:BLD_Failure}
  \vspace{-5mm}
\end{figure}

\section{Method}
In \cref{subsec:motivation}, we analyze the failure of blended sampling in few-step models and outline the motivations behind InverFill. The key component of our system is a one-step inversion network tailored for inpainting. We will present an overview on this network (\cref{sec:inversionnet}), followed by our proposed components in training pipeline (\cref{sec:losses,subsec:reblend,subsec:gaussreg,subsec:ladd,subsec:final}). Finally, we present our inpainting pipeline in \cref{Method: Inpaint_Pipeline}. \cref{fig:inversion_network} illustrates the training pipeline of our inversion network, while \cref{fig:inpainting_process} shows the full inpainting inference pipeline.

\subsection{Motivation}
\label{subsec:motivation}
\textbf{Blended Sampling Strategy for Few-Step Model.} While BLD is effective for multi-step diffusion models, applying it directly to few-step models significantly reduces inpainting quality. As shown in \cref{fig:BLD_Failure}, blending under few-step inference introduces semantic and stylistic inconsistencies between the generated and unmasked regions, yielding visible artifacts. This limitation originates from the initialized random Gaussian noise in the reverse process: multi-step models progressively refine this noise and adapt to the context of the unmasked regions within $I_m$, whereas few-step models make large ODE updates and lack sufficient refinement steps. When initialized from semantically distant noise, few coarse updates cannot correct the mismatch. Thus, effective few-step blending requires initializing $z_T$ semantically aligned with the unmasked regions of the image.

\noindent\textbf{Inversion for Image Inpainting.} A promising direction for mitigating semantic misalignment is diffusion inversion, which maps the unmasked image into the final noise latent $z_T$. However, existing inversion methods are iterative and introduce considerable overhead, contradicting the efficiency requirements of few-step sampling. A one-step inversion is critical for fast inference, as demonstrated by SwiftEdit~\cite{Nguyen_2025_CVPR}, which provides efficient and semantically coherent initialization. Nonetheless, directly applying SwiftEdit to inpainting is unsuitable for two reasons: (1) it is not explicitly designed for processing masked inputs, which causes information leakage during training, and (2) its training objectives do not enforce the inverted noise to follow the required Gaussian prior, resulting in distributional mismatch and degraded reconstructions.

To overcome these limitations, we introduce \textbf{InverFill}, a one-step inversion network designed for inpainting. InverFill (1) operates directly on masked images to produce semantically aligned initial noise latents, and (2) explicitly regularizes the inverted noise to match the Gaussian prior. Addressing both issues enables InverFill to achieve high-fidelity, coherent inpainting within the few-step regime.

\subsection{Masked Image Inversion Network}\label{sec:inversionnet}
\textbf{Problem Definition.}
Given a pretrained one-step text-to-image generator $\mathbf{G}$, we aim to develop a one-step inversion network $\mathbf{F_\theta}$ that is tailored for the inpainting purpose. Specifically, given a ground-truth image $I_{gt}$ and a masked image $I_m$ produced from $I_{gt}$ using a binary mask $M$, i.e., $I_m = I_{gt} \odot (1 - M)$, their image latents are ${z_0} = \mathcal{E}(I_{gt})$ and $z_0^{m} = \mathcal{E}(I_m)$, where $\mathcal{E}$ is the VAE encoder. We train $\mathbf{F_\theta}$ to map $z_0^m$ and text prompt $c$ to an inverted noise latent. The network is optimized so that passing this latent through $\mathbf{G}$ produces a predicted image latent $\hat{{z}}_0$ resembling the original latent ${z}_0$. The predicted noise latent should yield a reconstruction where (1) the background \textbf{faithfully preserves} the masked input, and (2) the generated region \textbf{harmonizes} with the background while remaining consistent with the text prompt and the unmasked content of $I_m$.

\noindent\textbf{Inversion Network Architecture.} Following \cite{Nguyen_2025_CVPR}, $\mathbf{F}_\theta$ shares the architecture of the one-step generator $\mathbf{G}$ and inherits its pretrained weights during as initialization.

\noindent\textbf{Masked Image Training.} To adapt our inversion network to masked image inputs, we leverage the one-step generator $\mathbf{G}$ to synthesize training image–mask–prompt triplets on the fly. Given a text prompt $c$ and random Gaussian noise $\epsilon \sim \mathcal{N}(0, I)$, $\mathbf{G}$ produces a ground-truth image latent ${z_0} = \mathbf{G}(\epsilon, c)$ and its corresponding image $I_{gt} = \mathcal{D}(z_0)$, where $\mathcal{D}$ is the VAE decoder. To ensure robustness and prevent overfitting to specific masks, we randomly sample a mask $M$ of diverse shapes and brush types and apply it to $I_{gt}$ to generate the masked image $I_m$. The subsequent masked image latent $z_{0}^m=\mathcal{E}(I_m)$ serves as input to our one-step inversion network $\mathbf{F}_\theta$, which predicts the inverted noise latent $\hat{z}_T$. In the following sections, we introduce our objective functions and describe how we optimize and integrate $\hat{z}_T$ to achieve a high-quality reconstruction of $z_{0}$.
\begin{figure}[t]
  \centering
  \includegraphics[width=1.0\linewidth]{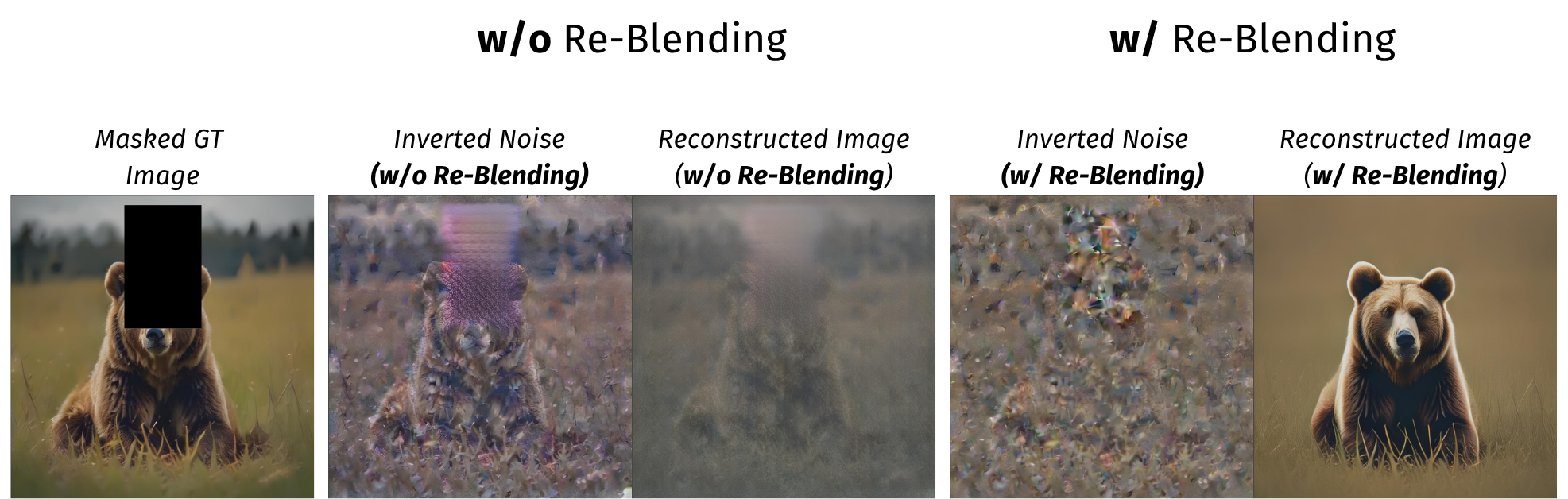}
  \caption{Effects of the proposed \textbf{Re-Blending operation} during training, without $\boldsymbol{\mathcal{L}_{\text{reg}}}$. \textit{(Zoom in for details)}}
  \label{fig:No_Blending}
  \vspace{-5mm}
\end{figure}

\noindent\subsection{Reconstruction Objectives}\label{sec:losses}
Similar to SwiftEdit~\cite{Nguyen_2025_CVPR}, we apply reconstruction losses in both the noise latent ($\mathcal{L}_{\text{noise}}$) and image latent ($\mathcal{L}_{\text{image}}$) spaces. Since our inversion network operates on a masked image $I_m$, applying $\mathcal{L}_{\text{noise}}$ over the entire predicted latent $\hat{z}_T$ is suboptimal: the regions of $z^m_0$ corresponding to the masked areas of $I_m$ contain no meaningful information, and penalizing these regions can hinder training. Therefore, we restrict $\mathcal{L}_{\text{noise}}$ to the unmasked regions. Our reconstruction objectives are formulated as follows:

\begin{equation}
    \mathcal{L}_{\text{noise}} = \| (1-m) \odot {\hat{z}}_T - (1-m) \odot \epsilon \|_2^2,
    \label{eq:noiseloss}
\end{equation}
\begin{equation}
    \mathcal{L}_{\text{image}} = \| {\hat{z}_0} - {z_0} \|_2^2
    \label{eq:imageloss}
\end{equation}
\begin{equation}
    \mathcal{L}_{\text{recons}} = \lambda_{\text{noise}}*\mathcal{L}_{\text{noise}} + \lambda_{\text{image}}*\mathcal{L}_{\text{image}}
    \label{eq:inversion}
\end{equation}

\noindent\subsection{Re-Blending Operation}\label{subsec:reblend}

Our inversion network maps the unmasked image content to a noise latent $\hat{z}_T$. In SwiftEdit~\cite{Nguyen_2025_CVPR}, $\mathcal{L}_{\text{noise}}$ ensures that the predicted noise latent $\hat{z}_T$ preserves details of the complete input image $I$ in the noise latent space. However, for inpainting tasks, our inversion network $\mathbf{F_\theta}$ only receives the incomplete masked image $I_m$ to predict $\hat{z}_T$. Consequently, our masked loss $\mathcal{L}_{\text{noise}}$ in \cref{eq:noiseloss} causes training bias towards the unmasked regions. This bias causes image-space structural patterns from $I_m$ to leak into $\hat{z}_T$, while regions corresponding to the mask exhibit low variance and artifacts. As a result, $\hat{z}_T$ deviates significantly from the Gaussian distribution expected by the diffusion model. During training, this distributional mismatch leads $\mathbf{G}$ to collapse when computing $\hat{z}_0 = \mathbf{G}(\hat{z}_T, c)$, producing the incoherent, artifact-filled outputs illustrated in \cref{fig:No_Blending}.

To address this, we introduce a \textbf{Re-Blending} operation. During training and inference, the masked regions of the predicted noise latent $\hat{z}_T$ are replaced with random Gaussian noise $\epsilon' \sim \mathcal{N}(0,I)$, partially restoring the latent to the expected distribution and recovering key semantic features, as shown in \cref{fig:No_Blending}. Following \cref{eq:blend}, generator $\mathbf{G}$ inputs the corrected latent $\hat{z}_T^{blend}$ to produce the final output ${\hat{z}_0}$.
\begin{equation}
    \hat{z}_T^{blend} = \hat{z}_T \odot (1-m) + \epsilon' \odot m, 
    \quad 
    {\hat{z}_0} = \mathbf{G}(\hat{z}_T^{blend}, c)
    \label{eq:blend}
\end{equation} where $m$ is latent-space mask downsampled from $M$.

\begin{figure}[t]
  \centering
  \includegraphics[width=1.0\linewidth]{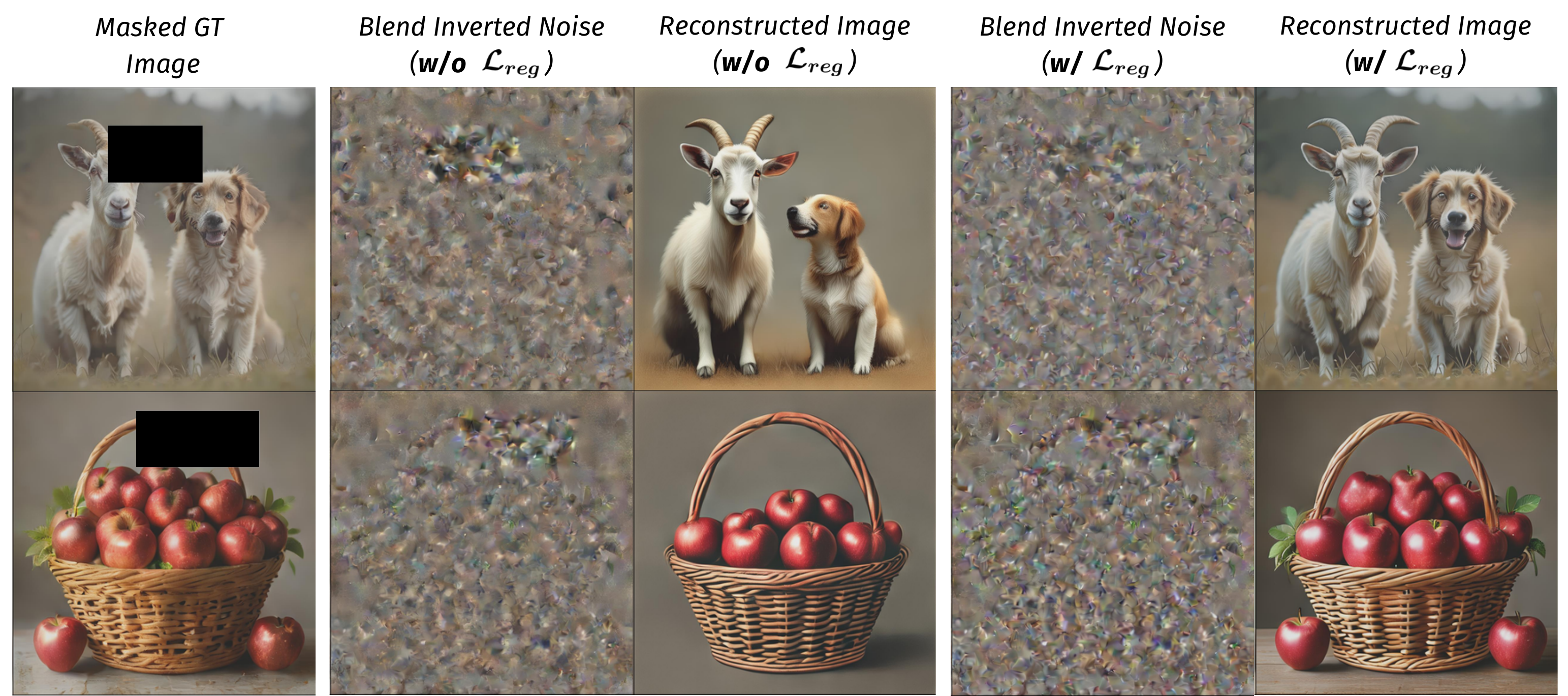}
  \caption{\textbf{Ablation Study on $\mathcal{L}_{\text{reg}}$}. Without the regularization loss, the model fails to preserve the background during reconstruction and produces blurred, low-detail outputs. With the loss, the background is well preserved and fine image details are restored.}
  \label{fig:Gauss_Reg_Need}
  \vspace{-5mm}
\end{figure}

\subsection{Gaussian Regularization}
\label{subsec:gaussreg}
As shown in \cref{fig:No_Blending}, Re-Blending mitigates information leakage and partially restores the Gaussian structure of $\hat{z}_T^{\text{blend}}$, but the predicted latent $\hat{z}_0$ recovers only key semantics. The output fails to preserve background, and the generated content within masked regions remains blurry and low-detail, as shown in \cref{fig:Gauss_Reg_Need}. Despite Re-Blending's partial correction, $\hat{z}_T^{\text{blend}}$ still deviates from the standard Gaussian expected by the generator. This occurs because $\mathcal{L}_\text{noise}$ in \cref{eq:noiseloss} focuses solely on the unmasked regions. While the image-space loss $\mathcal{L}_\text{image}$ in \cref{eq:imageloss} indirectly encourages the injected noise $\epsilon'$ to harmonize with $\hat{z}_T$ to better reconstruct image latent, it cannot fully enforce Gaussian consistency due to the lack of direct supervision with ground-truth Gaussian noise within the masked regions. As a result, the added $\epsilon'$ remains visually distinct from the background inverted noise in $\hat{z}_T$ (Column 2 in \cref{fig:Gauss_Reg_Need}), indicating that $\hat{z}_T^{\text{blend}}$ is still far from the target Gaussian distribution.

To address this issue, inspired by \cite{hwangmoment}, we introduce an additional Gaussian regularization term. This term explicitly encourages Gaussian distribution on the blended latent ${\hat{z}}_T^{\text{blend}}$ by matching its statistical moments with the theoretical moments of a standard Gaussian. Let $\mu_n$ be the $n$-th theoretical moment of a standard Gaussian. The moment-matching loss for the $n$-th moment is defined as:
\begin{equation}
    \mathcal{L}_n = \left\| 
     \left| \frac{1}{D} \sum_{k=1}^{D} \left( \hat{z}_T^{blend} \right)^n \right|^{\frac{1}{n}} 
    - \mu_n^{\frac{1}{n}} 
    \right\|,
\end{equation}
where $D = c \times h \times w$ is the total number of pixels of $\hat{z}_T^{blend}$. Our final regularization loss, $\mathcal{L}_{\text{reg}}$, is the sum of the losses for the first and second moments, corresponding to the mean and variance of Gaussian distribution: 
\begin{equation}
    \mathcal{L}_{\text{reg}} = \sum_{n \in \{1,2\}} \mathcal{L}_n
\end{equation}
As shown in \cref{fig:Gauss_Reg_Need}, Gaussian Regularization Loss during training helps ${\hat{z}}_T^{\text{blend}}$ better align with the Gaussian prior, enabling faithful reconstruction of the original image while preserving the background, as confirmed in \cref{tab:ablation_gaussian}.

\begin{table*}[t]
\centering
\footnotesize
\setlength{\tabcolsep}{4pt}
\caption{\textbf{Quantitative comparison of InverFill against few-step and multi-step diffusion inpainting baselines} on BrushBench and MagicBrush. NFEs denotes the number of function evaluations. $\uparrow$ indicates that higher is better, $\downarrow$ indicates that lower is better. IR scores are scaled by $10$, and HPSv2.1 scores are scaled by $10^2$.}
\label{tab:combined_final}
\begin{tabular}{@{}llc S[table-format=2.2] S[table-format=2.2] S[table-format=1.2] S[table-format=2.2] S[table-format=1.2] S[table-format=2.2] S[table-format=1.2] S[table-format=2.2] c@{}}
\toprule
\multirow{2}{*}{\textbf{Type}} & \multirow{2}{*}{\textbf{Method}} & \multirow{2}{*}{\textbf{NFEs}} & \multicolumn{4}{c}{\textbf{BrushBench}} & \multicolumn{4}{c}{\textbf{MagicBrush}} & \multirow{2}{*}{\textbf{Runtime}$\downarrow$} \\
\cmidrule(lr){4-7} \cmidrule(lr){8-11}
& & & {IR$\uparrow$} & {HPSv2.1$\uparrow$} & {AS$\uparrow$} & {CLIP$\uparrow$} & {IR$\uparrow$} & {HPSv2.1$\uparrow$} & {AS$\uparrow$} & {CLIP$\uparrow$} & (seconds)\\
\midrule
& SANA-Sprint 0.6B & 2 & {10.73} & {26.27} & {6.03} & {27.23} & {2.65} & {23.57} & {5.35} & {25.70} & {0.37} \\
\rowcolor{blue!15} \cellcolor{white} & SANA-Sprint 0.6B + InverFill & 2 & \textbf{{11.76}} & \textbf{{26.74}} & \textbf{{6.13}} & \textbf{{27.30}} & \textbf{{3.07}} & \textbf{{23.78}} & \textbf{{5.42}} & \textbf{{25.77}} & {0.43} \textcolor{red}{(+0.06)} \\
\cmidrule{2-12}
& SANA-Sprint 0.6B & 4 & {10.85} & {26.20} & {6.07} & {27.14} & {2.52} & {23.58} & {5.35} & {25.50} & {0.45} \\
\rowcolor{blue!15} \cellcolor{white} & SANA-Sprint 0.6B + InverFill & 4 & \textbf{{11.58}} & \textbf{{26.79}} & \textbf{{6.17}} & \textbf{{27.17}} & \textbf{{2.93}} & \textbf{{23.82}} & \textbf{{5.43}} & \textbf{{25.63}} & {0.51} \textcolor{red}{(+0.06)} \\
\cmidrule{2-12}
& SDXL Turbo & 4 & {11.60} & {27.26} & {6.06} & {27.21} & {3.42} & {24.09} & {5.45} & {25.76} & {0.66} \\
\rowcolor{blue!15} \cellcolor{white}& SDXL Turbo + InverFill & 4 & \textbf{{12.34}} & \textbf{{27.42}} & \textbf{{6.08}} & \textbf{{27.74}} & \textbf{{3.49}} & \textbf{{24.17}} & \textbf{{5.47}} & \textbf{{26.15}} & {0.70} \textcolor{red}{(+0.04)} \\
\cmidrule{2-12}
& SDXL Turbo + BrushNet & 4 & {12.30} & {27.37} & {6.30} & {27.70} & \textbf{{4.31}} & {23.87} & {5.28} & {25.97} & {0.70} \\
\rowcolor{blue!15} \cellcolor{white} \multirow{-9.5}{*}{\textbf{Few-step}}& SDXL Turbo + BrushNet + InverFill & 4 & \textbf{{12.34}} & \textbf{{27.42}} & \textbf{{6.33}} & \textbf{{27.75}} & \textbf{{4.31}} & \textbf{{24.06}} & \textbf{{5.30}} & \textbf{{26.23}} & {0.74} \textcolor{red}{(+0.04)} \\
\midrule
\rowcolor{gray!20} & SANA 0.6B & 20 & {11.86} & {26.15} & {6.47} & {27.68} & {3.79} & {23.06} & {5.56} & {26.28} & {1.18} \\
\rowcolor{gray!20}& HD-Painter & 30 & {12.56} & {27.28} & {6.60} & {27.62} & {3.70} & {23.55} & {5.73} & {26.22} & {23.45} \\
\rowcolor{gray!20}& SDXL-Inpainting & 30 & {12.90} & {28.03} & {6.67} & {27.34} & {4.02} & {23.08} & {5.59} & {25.85} & {3.35} \\
\rowcolor{gray!20}\multirow{-4}{*}{\textbf{Multi-step}}& SDXL + BrushNet & 30 & {13.00} & {27.39} & {6.56} & {27.73} & {4.05} & {23.23} & {5.54} & {25.95} & {4.31} \\
\bottomrule
\end{tabular}
\vspace{-2mm}
\end{table*}

\begin{figure*}[t]
  \centering
  \includegraphics[width=1.0\linewidth]{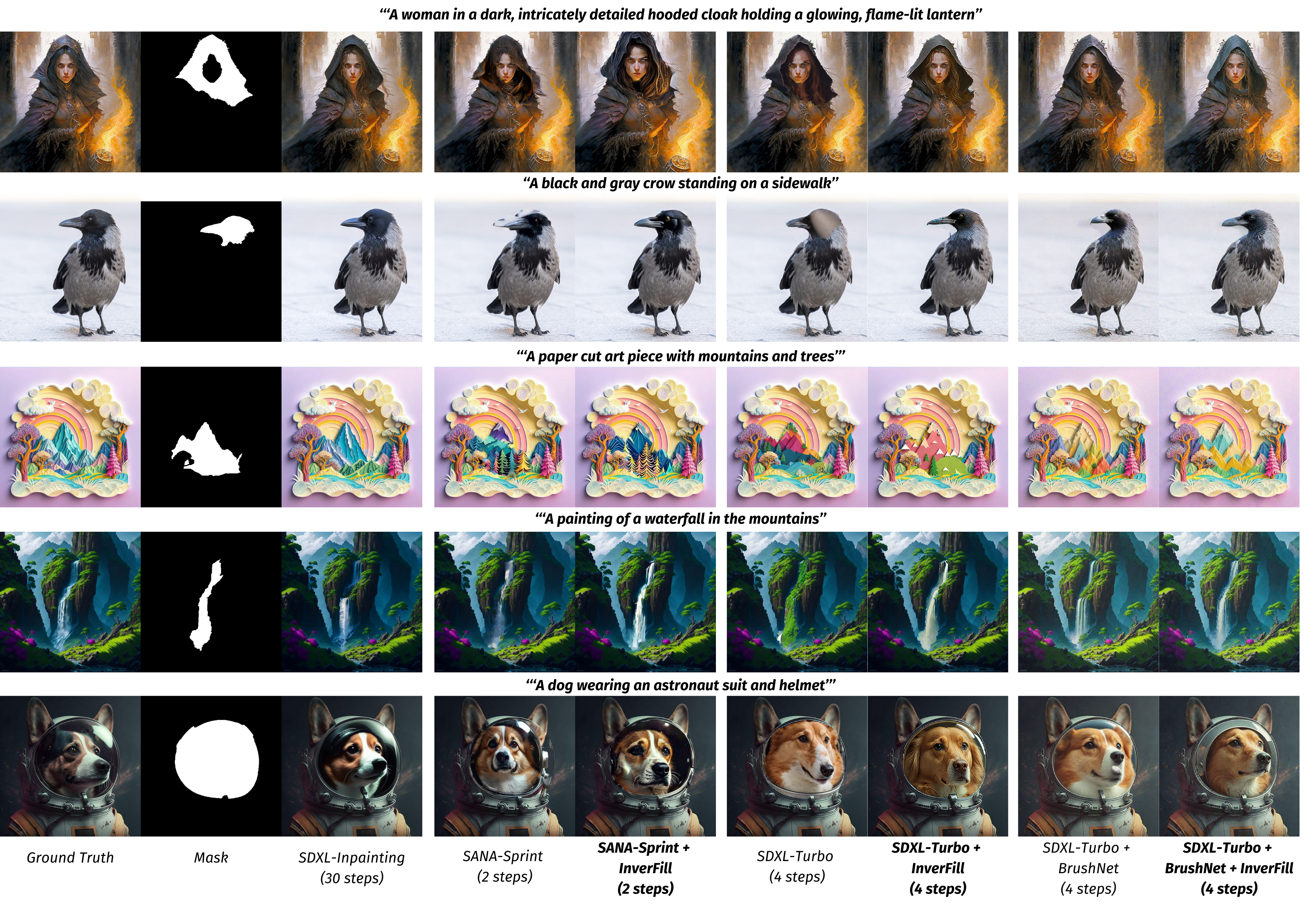}
    \caption{Our method achieves qualitative results comparable to multi-step SDXL-Inpainting and is on par with BrushNet (4 steps), as shown in Columns \textbf{7} and \textbf{8}. Notably, this performance is obtained using only text prompts during training, whereas competing methods rely on full text–image–mask supervision. Moreover, integrating our approach with BrushNet further enhances semantic coherence.} 
  \label{fig:Qualitative_Result}
\end{figure*}

\subsection{Improving Quality with Adversarial Loss}
\label{subsec:ladd}
Previous works~\cite{dmd, dmd2, sanasprint} show that adversarial losses during training improve visual quality. Following LADD~\cite{ladd}, we use the frozen teacher model to define a latent feature space for adversarial supervision, with multiple discriminator heads on intermediate layers for stable, efficient distillation. In our training, we treat the original image latent ${z}_0$ as real and the predicted image latent ${\hat{z}}_0$ as fake to train the inversion model and discriminator as follows:
\begin{equation}
    \mathcal{L}_{\text{adv}}^{G} (\theta) = 
    -\mathbb{E}_{{\hat{z}}_0, t} \left[
        \sum_k D_{\psi, k} \left( 
            G_{\text{pre}}\left({\hat{z}}_t, t, c \right)
        \right)
    \right]
    \label{eq:LG_adv}
\end{equation}
\begin{equation}
\begin{aligned}
    \mathcal{L}_{\text{adv}}^{D} & (\psi) = \mathbb{E}_{{z}_0, t} \left[ \sum_k \text{ReLU} \left( 1 - D_{\psi,k} \left( G_{\text{pre}} \left({z}_t, t, c \right) \right) \right) \right] \\
   & + \mathbb{E}_{{\hat{z}}_0, t} \left[ \sum_k \text{ReLU} \left( 1 + D_{\psi,k} \left( G_{\text{pre}} \left( {\hat{z}}_t, t, c \right) \right) \right) \right] \label{eq:LD_adv_2}
\end{aligned}
\end{equation}
\noindent where $z_t$, ${\hat{z}}_t$ are noisy versions of original image latent $z_0$ and predicted image latent ${\hat{z}}_0$ at timesteps $t$. $G_{\text{pre}}$ denotes a frozen multi-step teacher model. $D_{\psi,k}$ denotes discriminator heads at the $k$-th intermediate layers of $G_{\text{pre}}$.

\subsection{Final Objectives}
\label{subsec:final}
Our final training objective for $\mathbf{F_\theta}$ is defined as follows:
\begin{equation}
    \mathcal{L}_{\text{final}} = \lambda_{\text{recons}}*\mathcal{L}_{\text{recons}} + \lambda_{\text{reg}}*\mathcal{L}_{\text{reg}} + 
    \lambda_{\text{adv}}*\mathcal{L}_{\text{adv}}
    \label{eq:final_eq}
\end{equation}

\subsection{Inpainting Pipeline} \label{Method: Inpaint_Pipeline}
\cref{fig:inpainting_process} illustrates our inpainting pipeline, which closely follows the blended sampling strategy described in \cref{subsec:blendsample}. However, instead of initializing with random Gaussian noise, we employ our trained inversion model $\mathbf{F_\theta}$ to predict the inverted noise latent $\hat{z}_T$ and obtain the blended latent $\hat{z}_T^{blend}$ using \cref{eq:blend}. This blended latent $\hat{z}_T^{blend}$ serves as the Gaussian noise input to the inpainting process.

\section{Experiments}
\label{sec:exp}
\subsection{Training Details}
We train InverFill on Sana-Sprint 0.6B~\cite{sanasprint} and SDXL-Turbo~\cite{add}, which represent two common diffusion architectures: DiT and UNet. All training is performed on four NVIDIA A100 40GB GPUs for 8-10 hours. During training, we randomly sample text prompts from BrushData~\cite{brushnet} and MSCOCO~\cite{mscoco}. We use a total batch size of 32 and a learning rate of $1\times10^{-5}$ with AdamW optimizer.

\subsection{Evaluation Setup}
\noindent\textbf{Dataset.} We perform evaluation on inpainting BrushBench~\cite{brushnet}, with 600 images and annotated masks, and image editing MagicBrush~\cite{magicbrush} benchmark. For inpainting, we adapt MagicBrush’s 535-image test set using its captions and masks. Each image includes multiple segmentation and random masks, providing a diverse and challenging evaluation for inpainting performance. All experiments and evaluations were performed using $1024^2$ resolution.

\noindent\textbf{Evaluation Metrics.} We evaluate our results from two criteria: image generation quality and text alignment.
\begin{itemize}
\item \textit{Image Generation Quality.} We use three human-aligned metrics: ImageReward (IR) \cite{imagereward}, HPS v2 (HPS) \cite{hps}, and Aesthetic Score (AS) \cite{aesthetics}. IR and HPS are reward models trained on large-scale human preference data, while AS is a linear model trained to predict perceptual quality.
\item \textit{Text Alignment.} We measure text–image alignment using CLIP Similarity (CLIP) \cite{clip}, which quantifies how well the inpainted images match their prompts.
\end{itemize}

\noindent\textbf{Baselines.} We evaluate InverFill on state-of-the-art few-step text-to-image diffusion models, SANA-Sprint 0.6B~\cite{sanasprint} and SDXL-Turbo~\cite{add}, using the blended sampling strategy in \cref{subsec:blendsample} for inpainting. We report results using 2- and 4-step NFE settings for SANA-Sprint and 4-step for SDXL-Turbo. Following~\cite{turbofill}, we integrate SDXL-Turbo~\cite{sdxl} with BrushNet~\cite{brushnet} to evaluate InverFill using few-step specialized inpainting models that do not rely on the blended sampling strategy. For reference, we report results from multi-step models, including Sana 0.6B~\cite{sana}, HD-Painter~\cite{manukyan2023hd}, SDXL-Inpainting \cite{sdxl} and SDXL with BrushNet \cite{brushnet}.

\begin{table}[t]
\centering
\footnotesize
\setlength{\tabcolsep}{2pt}
\caption{Effects of $\boldsymbol{\mathcal{L}_{\text{reg}}}$ on SANA-Sprint 0.6B~\cite{sanasprint} with 2 NFEs on BrushBench~\cite{brushnet}. All models were evaluated at 5000 iterations.}
\begin{tabular}{lcccc}
\toprule
\multirow{1}{*}{\textbf{Method}} 
& \text{IR}$\uparrow$ & \text{HPSv2.1}$\uparrow$ & \text{AS}$\uparrow$ & \text{CLIP}$\uparrow$ \\
\midrule
w/o $\mathcal{L}_{reg}$ & 11.22 & 25.49 & 6.06 & 27.26  \\
\midrule
w/ $\mathcal{L}_{reg}$ & \textbf{11.51} & \textbf{26.03} & \textbf{6.10} & \textbf{27.28}
  
\\
\bottomrule
\end{tabular}
\label{tab:ablation_gaussian}
\end{table}

\subsection{Quantitative Results}
As shown in \cref{tab:combined_final}, InverFill consistently improves performance across few-step diffusion settings. When integrated with SANA-Sprint and SDXL-Turbo under blended sampling, InverFill boosts all metrics on BrushBench and MagicBrush. For example, SANA-Sprint (2 NFEs) + InverFill raises IR from 10.73 to \textbf{11.76} on BrushBench and 2.65 to \textbf{3.07} on MagicBrush. InverFill also strengthens specialized inpainting model. In BrushNet + InverFill (4 NFEs), IR improves from 12.30 to \textbf{12.34} and HPS from 27.37 to \textbf{27.42}. Regarding text alignment, InverFill achieves substantial gains in CLIP scores. Notably, InverFill-equipped few-step models match or surpass multi-step methods while remaining efficient; SDXL-Turbo + InverFill (4 NFEs) outperforms HD-Painter (30 NFEs) on key metrics. Despite these gains, InverFill introduces extremely minimal overhead, only \textbf{0.06s} on SANA-Sprint and \textbf{0.04s} on SDXL.

\subsection{Qualitative Results} 
\cref{fig:Qualitative_Result} shows that integrating InverFill improves coherence and background harmonization. Without using real images, InverFill achieves quality comparable to BrushNet (4-step SDXL-Turbo), which relies on an inpainting dataset of real images~\cite{brushnet}. Moreover, combining InverFill with the BrushNet + SDXL-Turbo pipeline further boosts semantic quality, indicating that InverFill can also strengthen specialized few-step inpainting systems.

\subsection{Enhanced Caption for BrushBench}
\label{subsec:enhanced}
\noindent\textbf{Motivation.}
A limitation of BrushBench~\cite{brushnet} is its reliance on simple, short prompts, which limits evaluation of text understanding and compositional generation. Modern models, SDXL~\cite{sdxl} with dual text encoders and SANA-Sprint~\cite{sanasprint} with Gemma-2~\cite{riviere2024gemma}, are built for more context-heavy prompts. Therefore, we expand BrushBench captions using Qwen3~\cite{yang2025qwen3} with detailed foreground and background descriptions, enabling more comprehensive evaluation of text alignment and visual coherence in inpainting.

\noindent\textbf{Quantitative Results.} \cref{tab:enhanced_table} shows that InverFill remains effective under detailed, complex prompts, improving all baselines and demonstrating robustness in text-rich settings. For SANA-Sprint, CLIP gains exceed those with simple prompts in \cref{tab:combined_final}, indicating stronger visual–text alignment and better use of large encoders like Gemma-2.

\begin{table}[t]
\centering
\scriptsize
\setlength{\tabcolsep}{2pt}
\caption{Quantitative comparison of InverFill against few-step and multi-step diffusion inpainting baselines on \textbf{BrushBench with enhanced prompts}. $\uparrow$ indicates that higher is better.}
\begin{tabular}{@{}llc S[table-format=2.2] S[table-format=2.2] S[table-format=1.2] S[table-format=2.2]@{}}
\toprule
& \textbf{Method} & \textbf{NFEs} & \text{IR$_{\times 10}$}$\uparrow$ & \text{HPS$_{\times 10^2}$}$\uparrow$ & \text{AS}$\uparrow$ & \text{CLIP}$\uparrow$ \\
\midrule
& SANA-Sprint 0.6B & 2 & {7.53} & {25.81} & {6.09} & {28.10} \\
\rowcolor{blue!15} \cellcolor{white} &  SANA-Sprint 0.6B + InverFill & 2 & \textbf{8.55} & \textbf{26.49} & \textbf{6.19} & \textbf{28.26} \\
\addlinespace[0.5em]
& SANA-Sprint 0.6B & 4 & {7.32} & {25.81} & {6.11} & {28.15} \\
\rowcolor{blue!15} \cellcolor{white} &  SANA-Sprint 0.6B + InverFill & 4 & \textbf{8.54} & \textbf{26.50} & \textbf{6.21} & \textbf{28.29} \\
\addlinespace[0.5em]
& SDXL Turbo & 4 & {8.09} & {26.85} & {6.12} & {28.25} \\
\rowcolor{blue!15} \cellcolor{white} & SDXL Turbo + InverFill & 4 & \textbf{9.00} & \textbf{27.11} & \textbf{6.12} & \textbf{28.70} \\
\addlinespace[0.5em]
& SDXL Turbo + BrushNet & 4 & {9.35} & {27.22} & {6.06} & {28.87} \\
\rowcolor{blue!15} \cellcolor{white} \parbox[t]{2mm}{\multirow{-9}{*}{\rotatebox[origin=c]{90}{\textbf{Few-step}}}} &  SDXL Turbo + BrushNet + InverFill & 4 & \textbf{9.53} & \textbf{27.37} & \textbf{6.10} & \textbf{28.91} \\
\midrule\midrule
 \rowcolor{gray!20} & SANA 0.6B & 20 & {9.34} & {26.77} & {6.24} & {28.63} \\
 \rowcolor{gray!20} & HD-Painter & 30 & {9.60} & {27.83} & {6.35} & {28.63} \\
 \rowcolor{gray!20} & SDXL-Inpainting & 30 & {9.90} & {27.43} & {6.38} & {28.38} \\
 \rowcolor{gray!20} \parbox[t]{2mm}{\multirow{-4}{*}{\rotatebox[origin=c]{90}{\textbf{Multi-step}}}} & SDXL + BrushNet & 30 & {10.41} & {28.11} & {6.34} & {28.76} \\
\bottomrule
\end{tabular}
\label{tab:enhanced_table}
\vspace{-5mm}
\end{table}
\section{Societal Impacts}
Our work aims to provide a practical tool for creative professionals, facilitating tasks such as photo restoration and object removal. We acknowledge that realistic image manipulation technologies can be misused to generate deceptive content. To mitigate such risks, we advocate for the parallel development of detection methods \cite{suma,erase,nguyen2025cgce,Vu_2026_CVPR} for AI-manipulated media and encourage the responsible use of these technologies.

\section{Conclusion}
In this work, we introduce \textbf{InverFill}, a lightning-fast one-step inversion network explicitly designed for image inpainting that enhances existing few-step inpainting methods. Extensive experiments show that InverFill produces high-quality inpainting results while adding as few as \textbf{0.06 seconds} of overhead.

\clearpage
{
    \small
    \bibliographystyle{ieeenat_fullname}
    \bibliography{main}
}

\clearpage
\newcommand{\ffhq}[1]{\textcolor{red}{#1}}
\newcommand{\divk}[1]{\textcolor{blue}{#1}}

\setcounter{page}{1}
\maketitlesupplementary
\noindent We first present ablations on the loss weights in \cref{sec:weight}. \cref{sec:sds} compares our method with other regularization techniques, while \cref{sec:inversion} evaluates alternative inversion methods. Additional ablations for our proposed components are in \cref{sec:top}. \cref{sec:morequal} includes qualitative comparisons.

\noindent\textbf{Note:} All experiments in both the main paper and supplementary use $1024^2$ resolution. For all supplementary results, we use BrushBench~\cite{brushnet} with its original captions.

\section{Loss Weight Ablations}
\label{sec:weight}
We evaluate the impact of reconstruction weights $\lambda_{\text{noise}}$ and $\lambda_{\text{image}}$ in $\mathcal{L}_{\text{recons}}$ (\cref{sec:losses}), along with the Gaussian regularization $\lambda_{\text{reg}}$ (\cref{subsec:gaussreg}) and adversarial weights $\lambda_{\text{adv}}$ (\cref{subsec:ladd}), using SANA-Sprint 0.6B~\cite{sanasprint}. All experiments use the Re-Blending operation (\cref{subsec:reblend}) during training and inference. For LADD adversarial loss, the discriminator learning rate is set to $1\times10^{-6}$. Detailed results are provided in \cref{tab:weight}.

\begin{table}[H]
\centering
\footnotesize
\captionsetup{font=small,skip=0pt}
\caption{Ablation study on key hyperparameters for each component. The best setting from each block is propagated to the next.}
\label{tab:weight}

\resizebox{0.98\linewidth}{!}{
\begin{tabular}{c c c c c c c c c}
\toprule
Method & $\lambda_{\text{noise}}$ & $\lambda_{\text{image}}$ & $\lambda_{\text{reg}}$ & $\lambda_{\text{adv}}$ & \text{IR}$\uparrow$ & \text{HPSv2.1}$\uparrow$ & \text{AS}$\uparrow$ & \text{CLIP}$\uparrow$ \\
\midrule

\multirow{3}{*}{\cref{sec:losses}} 
& \cellcolor{cyan!15} 2.0 & \cellcolor{cyan!15} 1.0 & 0 & 0 & 11.20 & \textbf{25.50} & 6.02 & 27.23 \\
& \cellcolor{cyan!15} 1.0 & \cellcolor{cyan!15} 2.0 & 0 & 0 & 11.02 & 25.48 & 6.04 & 27.18 \\
& \cellcolor{cyan!15} \textbf{1.0} & \cellcolor{cyan!15} \textbf{1.0} & 0 & 0 & \textbf{11.22} & \underline{25.49} & \textbf{6.06} & \textbf{27.26} \\
\midrule

\multirow{4}{*}{\cref{subsec:gaussreg}}
& 1.0 & 1.0 & \cellcolor{yellow!15} 0.25 & 0 & 11.23 & 25.36 & 6.07 & 27.26 \\
& 1.0 & 1.0 & \cellcolor{yellow!15} \textbf{0.5} & 0 & \textbf{11.51} & \textbf{26.03} & \textbf{6.10} & \underline{27.28} \\
& 1.0 & 1.0 & \cellcolor{yellow!15} 1.0 & 0 & 11.47 & 25.39 & 6.08 & 27.27 \\
& 1.0 & 1.0 & \cellcolor{yellow!15} 2.0 & 0 & 11.14 & 25.37 & 6.07 & \textbf{27.30} \\
\midrule

\multirow{3}{*}{\cref{subsec:ladd}}
& 1.0 & 1.0 & 0.5 & \cellcolor{green!15} 0.25 & 11.68 & 26.17 & 6.12 & 27.29 \\
& 1.0 & 1.0 & 0.5 & \cellcolor{green!15} \textbf{0.5} & \textbf{11.76} & \textbf{26.74} & \textbf{6.13} & \textbf{27.30} \\
& 1.0 & 1.0 & 0.5 & \cellcolor{green!15} 1.0 & 11.71 & 26.42 & 6.13 & 27.30 \\

\bottomrule
\end{tabular}}
\end{table}
\cref{tab:weight} summarizes the ablation results on the loss-weight components. Based on this study, we use the final weights $\lambda_{\text{noise}}=1.0$, $\lambda_{\text{image}}=1.0$, $\lambda_{\text{reg}}=0.5$, and $\lambda_{\text{adv}}=0.5$ for all experiments reported in \cref{tab:combined_final,tab:enhanced_table}.

\section{Comparison with Regularization Loss in SwiftEdit}
\label{sec:sds}
\begin{figure}[t]
  \centering
  \includegraphics[width=1.0\linewidth]{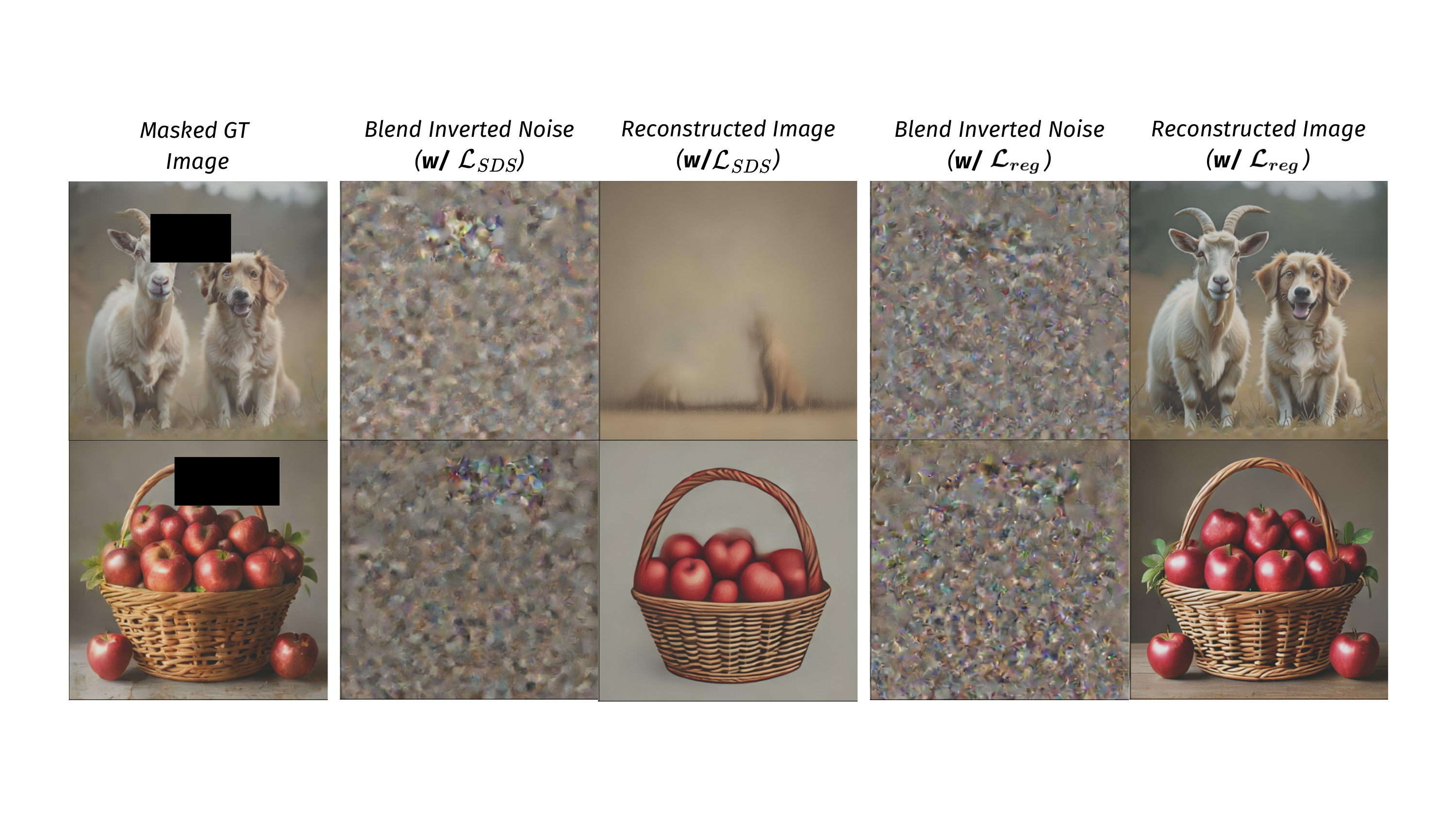}
   \caption{
        \textbf{Qualitative comparison between our proposed regularization loss ($\mathcal{L}_{{reg}}$) and the Score Distillation Sampling (SDS) loss ($\mathcal{L}_{\text{SDS}}$) from SwiftEdit \cite{Nguyen_2025_CVPR}.} 
        This visualization shows that our $\mathcal{L}_{reg}$ is crucial for preserving the original image content, while using $\mathcal{L}_{\text{SDS}}$ leads to significant information loss and poor reconstruction.
    }
  \label{fig:SDS_Ablation}
  \vspace{-5mm}
\end{figure}

We perform an ablation to compare our regularization loss $\mathcal{L}_{reg}$ with the Score Distillation Sampling loss $\mathcal{L}_{\text{SDS}}$ used in SwiftEdit ~\cite{Nguyen_2025_CVPR}. As shown in \cref{tab:Other_reg}, applying $\mathcal{L}_{reg}$ consistently outperforms $\mathcal{L}_{\text{SDS}}$ across all metrics (IR, HPS, AS, and CLIP), demonstrating its effectiveness in preserving image fidelity. \cref{fig:SDS_Ablation} illustrates the qualitative difference between the two losses. With the SDS-based loss, the reconstruction collapses, as the inverted noise is over-regularized and loses the semantic structure of the original image, producing blurry and unrecognizable results. In contrast, our Gaussian regularization loss $\mathcal{L}_{reg}$ preserves the semantic content and enables high-fidelity reconstruction from the inverted noise.

\begin{table}[t]
\centering
\footnotesize
\setlength{\tabcolsep}{2pt}
\caption{Quantitative comparison between the SDS loss $\mathcal{L}_{\text{SDS}}$ and our Gaussian regularization loss $\mathcal{L}_{reg}$. For a fair evaluation, both methods are tested on SANA-Sprint 0.6B using 2 NFEs.}
\begin{tabular}{lcccc}
\toprule
\multirow{1}{*}{\textbf{Method}} 
& \text{IR}$\uparrow$ & \text{HPSv2.1}$\uparrow$ & \text{AS}$\uparrow$ & \text{CLIP}$\uparrow$ \\
\midrule
$\mathcal{L}_{\text{SDS}}$ \cite{Nguyen_2025_CVPR} & 11.29 & 25.31 & 6.08 & 27.25  \\
\midrule
$\mathcal{L}_{reg}$ (Ours) & \textbf{11.51} & \textbf{26.03} & \textbf{6.10} & \textbf{27.28} \\
\bottomrule
\end{tabular}
\vspace{-3mm}
\label{tab:Other_reg}
\end{table}

\begin{table*}[t]
\centering
\setlength{\tabcolsep}{5pt}
\small
\begin{tabular}{lcccccccc} 
\toprule
\multirow{1}{*}{\textbf{Method}} & \text{IR}$\uparrow$ & \text{HPSv2.1}$\uparrow$ & \text{AS}$\uparrow$ & \text{CLIP}$\uparrow$ & \text{Runtime} (seconds)$\downarrow$ \\ 
\midrule
DDIMInv (\textbf{w/o} Blending) (50 steps) + SDXL-Turbo  & 4.06 & 21.93 & 5.30 & 26.47 & 4.18 \\
DDIMInv (\textbf{w/} Blending) (50 steps) + SDXL-Turbo  & 12.07 & 27.19 & 6.04 & 27.39 & 4.32 \\
\midrule
\textbf{InverFill (Ours) + SDXL-Turbo} & \textbf{12.34} & \textbf{27.42} & \textbf{6.08} & \textbf{27.74} & \textbf{0.74} \\
\bottomrule
\end{tabular}
\caption{Quantitative comparison of one-step InverFill versus the 50-step DDIM inversion baseline on BrushBench. For a fair comparison, we run SDXL-Turbo with 4 NFEs.}
\label{tab:Other_Inversion}
\vspace{-3mm}
\end{table*}

\section{Ablation of Proposed Components}
\label{sec:top}
To better understand the contribution of each part in our framework, we conducted an ablation study on both SANA-Sprint 0.6B (\cref{tab:top_up}) and SDXL-Turbo (\cref{tab:top_up_sdxl}). We established a baseline for comparison by training a model with the reconstruction loss from \cref{sec:losses}, using the masked image as input. From this starting point, we then incrementally added our proposed components: Re-Blending (\cref{subsec:reblend}), Gaussian Regularization (\cref{subsec:gaussreg}), and the LADD adversarial loss (\cref{subsec:ladd}).

Our results show that each component contributes incremental gains in performance. As shown in \cref{tab:top_up}, introducing the Re-Blending operation increases the IR score from 7.93 to 11.22. The further addition of Gaussian Regularization expands this improvement, and incorporating the LADD adversarial loss leads to the highest scores, with an IR of 11.76 and an HPS of 26.74. A similar pattern of improvement is also noted in the experiments with SDXL-Turbo (\cref{tab:top_up_sdxl}). This evaluation suggests that all three components contribute effectively, collectively leading to the performance of the full InverFill model.

\begin{table}[H]
    \centering
    \scriptsize
    \captionsetup{font=small}
        \caption{Ablation of components on SANA-Sprint 0.6B (2 NFEs).}
        \begin{tabular}{l c c c c} 
            \toprule
            \textbf{Method} & \text{{IR$_{\times 10}$}}$\uparrow$ & \text{HPS}$_{\times 10 ^{2}}$$\uparrow$ & \text{AS}$\uparrow$ & \text{CLIP}$\uparrow$ \\
            \midrule
            \rowcolor{gray!15} Baseline \cite{Nguyen_2025_CVPR} & 7.93 & 24.79 & 5.96 & 26.40 \\
            \midrule
            \multicolumn{5}{@{}c@{}}{\textbf{\textit{InverFill}}} \\ 
            \midrule
            + Re-Blending (\cref{subsec:reblend}) & 11.22 & 25.49 & 6.06 & 27.26 \\
            + Gaussian Reg. (\cref{subsec:gaussreg}) & 11.51 & 26.03 & 6.10 & 27.28 \\
            + LADD (\cref{subsec:ladd}) & \textbf{11.76} & \textbf{26.74} & \textbf{6.13} & \textbf{27.30} \\
            \bottomrule
        \end{tabular}
    \label{tab:top_up}
    \vspace{-5mm}
\end{table}

\begin{table}[H]
    \centering
    \scriptsize
    \captionsetup{font=small}
        \caption{Ablation of components on SDXL-Turbo (4 NFEs).}
        \begin{tabular}{l c c c c}
            \toprule
            \textbf{Method} & \text{IR$_{\times 10}$}$\uparrow$ & \text{HPS}$_{\times 10 ^{2}}$$\uparrow$ & \text{AS}$\uparrow$ & \text{CLIP}$\uparrow$ \\
            \midrule
            \rowcolor{gray!15} Baseline \cite{Nguyen_2025_CVPR} & 10.60 & 25.44 & 6.03 & 26.63 \\
            \midrule
            \multicolumn{5}{@{}c@{}}{\textbf{\textit{InverFill}}} \\
            \midrule
            + Re-Blending (\cref{subsec:reblend}) & 11.29 & 26.16 &  6.03 & 27.23 \\
            + Gaussian Reg. (\cref{subsec:gaussreg}) & 12.10	 & 27.12 & 6.06 & 27.64 \\
            + LADD (\cref{subsec:ladd}) & \textbf{{12.34}} & \textbf{{27.42}} & \textbf{{6.08}} & \textbf{{27.74}} \\
            \bottomrule
        \end{tabular}
    \label{tab:top_up_sdxl}
\end{table}

\section{Other Inversion Approaches}
\label{sec:inversion}
\begin{figure}[t]
  \centering
  \includegraphics[width=1.0\linewidth]{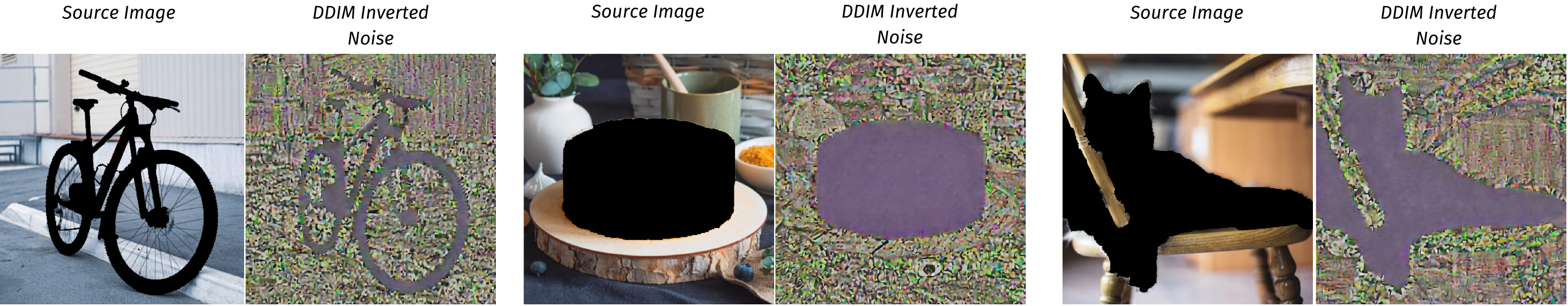}
   \caption{Visualization of DDIM Inversion results. The masked regions are not encoded, producing smooth, null-like areas in the inverted noise and causing loss of content.}
  \label{fig:DDIM_Noise}
\end{figure}

We quantitatively compare InverFill with a 50-step DDIM inversion process, using SDXL for inversion and SDXL-Turbo blended sampling for inpainting. Based on \cref{fig:DDIM_Noise}, directly applying DDIM inversion to a masked image fails to encode the masked regions, producing smooth, gray, null-like structures in those areas. This loss of content significantly degrades performance, as reflected in the low scores reported in the first row of \cref{tab:Other_Inversion}.

\begin{figure}[t]
  \centering
  \includegraphics[width=1.0\linewidth]{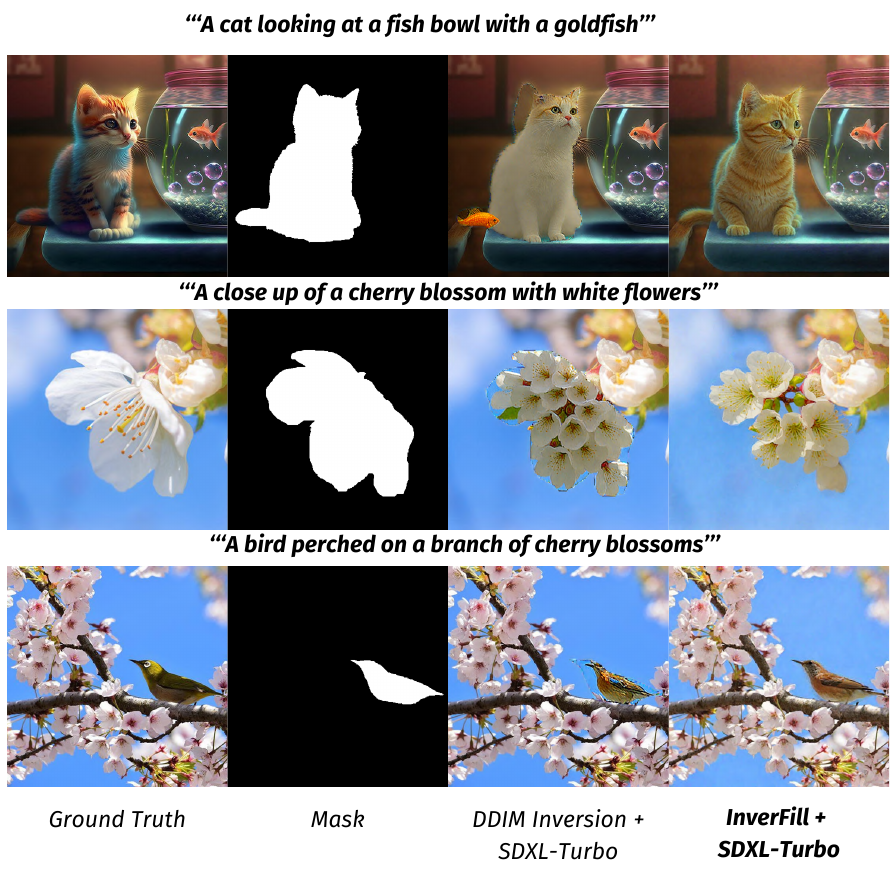}
   \caption{Qualitative comparison between InverFill and DDIM Inversion. InverFill achieves substantially better scene harmonization and semantic consistency.}
  \label{fig:DDIM_Qual}
  \vspace{-5mm}
\end{figure}

Next, we apply our proposed Re-Blending operation (\cref{subsec:reblend}) to the DDIM-inverted noise. While this fills the previously null-like regions, the resulting model (Row 2 in \cref{tab:Other_Inversion}) still struggles with scene harmonization. In contrast, our one-step InverFill method (Row 3) achieves higher scores across all metrics and is significantly more efficient, running in just \textbf{0.74 seconds}, nearly \textbf{six times faster} than the 50-step DDIM process (4.32 seconds). Combined with the improved qualitative harmonization in \cref{fig:DDIM_Qual}, these results demonstrate that InverFill is both substantially more effective and practical.

\begin{table*}[!t]
\centering
\caption{\textbf{Quantitative results on FFHQ, DIV2K, and BrushBench.} \ffhq{\textbf{Red}}, \divk{\textbf{Blue}}, and \textbf{Black} denote scores on \ffhq{\textbf{FFHQ}}, \divk{\textbf{DIV2K}}, and \textbf{BrushBench}, respectively.}
\label{tab:FFHQ_Div2k}
\vspace{-1mm}
\resizebox{\textwidth}{!}{%
\small
\setlength{\tabcolsep}{3.5pt}
\begin{tabular}{@{}lcccccccccc}
\toprule
\textbf{Method} & \textbf{NFEs}
& \multicolumn{7}{c}{\ffhq{\textbf{FFHQ}} / \divk{\textbf{DIV2K}} / \textbf{BrushBench}} \\
\cmidrule(lr){3-9}
& 
& FID$\downarrow$
& \text{IR$_{\times 10}$}$\uparrow$
& \text{HPS$_{\times 10^2}$}$\uparrow$
& AS$\uparrow$
& CLIP$\uparrow$
& LPIPS$\downarrow$
& SSIM$\uparrow$\\
\midrule

SANA-Sprint 0.6B & 2
& \ffhq{27.12}
& \ffhq{2.81} / \divk{5.22}
& \ffhq{22.42} / \divk{26.78}
& \ffhq{5.12} / \divk{5.77}
& \ffhq{23.55} / \divk{28.26}
& \ffhq{0.184} / \divk{0.193} / 0.144
& \ffhq{0.704} / \divk{0.572} / 0.769\\

\rowcolor{blue!15}SANA-Sprint 0.6B + InverFill & 2
& \ffhq{\textbf{26.53}}
& \ffhq{\textbf{5.27}} / \divk{\textbf{5.87}}
& \ffhq{\textbf{23.26}} / \divk{\textbf{27.17}}
& \ffhq{\textbf{5.31}} / \divk{\textbf{5.89}}
& \ffhq{\textbf{23.65}} / \divk{\textbf{28.43}}
& \ffhq{\textbf{0.172}} / \divk{\textbf{0.182}} / \textbf{0.138}
& \ffhq{\textbf{0.719}} / \divk{\textbf{0.575}} / \textbf{0.771}\\

\midrule
SANA-Sprint 0.6B & 4
& \ffhq{27.32}
& \ffhq{2.66} / \divk{5.25}
& \ffhq{22.50} / \divk{26.83}
& \ffhq{5.17} / \divk{5.79}
& \ffhq{23.84} / \divk{28.31}
& \ffhq{0.184} / \divk{0.192} / 0.140
& \ffhq{0.706} / \divk{0.573} / 0.774
\\

\rowcolor{blue!15}SANA-Sprint 0.6B + InverFill & 4
& \ffhq{\textbf{26.42}}
& \ffhq{\textbf{5.27}} / \divk{\textbf{5.83}}
& \ffhq{\textbf{23.32}} / \divk{\textbf{27.15}}
& \ffhq{\textbf{5.37}} / \divk{\textbf{5.92}}
& \ffhq{\textbf{23.88}} / \divk{\textbf{28.38}}
& \ffhq{\textbf{0.169}} / \divk{\textbf{0.181}} / \textbf{0.134}
& \ffhq{\textbf{0.708}} / \divk{\textbf{0.574}} / 0.774 
\\

\midrule
SDXL Turbo & 4
& \ffhq{26.32}
& \ffhq{7.37} / \divk{4.71}
& \ffhq{25.73} / \divk{26.81}
& \ffhq{5.67} / \divk{5.92}
& \ffhq{25.24} / \divk{28.21}
& \ffhq{0.269} / \divk{0.292} / 0.139
& \ffhq{0.626} / \divk{0.454} / 0.813
\\

\rowcolor{blue!15}SDXL Turbo + InverFill & 4
& \ffhq{\textbf{25.90}}
& \ffhq{\textbf{8.35}} / \divk{\textbf{5.27}}
& \ffhq{\textbf{26.14}} / \divk{\textbf{27.03}}
& \ffhq{\textbf{5.76}} / \divk{\textbf{5.95}}
& \ffhq{\textbf{25.29}} / \divk{\textbf{28.25}}
& \ffhq{\textbf{0.262}} / \divk{\textbf{0.287}} / \textbf{0.133}
& \ffhq{\textbf{0.655}} / \divk{\textbf{0.455}} / \textbf{0.815}\\

\midrule
SDXL Turbo + BrushNet & 4
& \ffhq{25.55}
& \ffhq{7.86} / \divk{5.11}
& \ffhq{25.05} / \divk{26.05}
& \ffhq{5.53} / \divk{5.76}
& \ffhq{24.72} / \divk{\textbf{28.41}}
& \ffhq{\textbf{0.204}} / \divk{0.469} / 0.185
& \ffhq{\textbf{0.728}} / \divk{0.292} / 0.755
\\

\rowcolor{blue!15}SDXL Turbo + BrushNet + InverFill & 4
& \ffhq{\textbf{25.49}}
& \ffhq{\textbf{7.91}} / \divk{\textbf{5.17}}
& \ffhq{\textbf{25.17}} / \divk{\textbf{26.18}}
& \ffhq{\textbf{5.55}} / \divk{\textbf{5.79}}
& \ffhq{\textbf{24.85}} / \divk{28.39}
& \ffhq{0.206} / \divk{0.469} / \textbf{0.178}
& \ffhq{0.727} / \divk{\textbf{0.293}} / \textbf{0.757}
\\

\bottomrule
\end{tabular}
}
\vspace{-4mm}
\end{table*}

\noindent\textbf{Datasets.} For FFHQ, we sample 10K images. For DIV2K, we use 900 images from the training and validation sets. Following the same settings in \cref{subsec:enhanced}, prompts are generated using Qwen-3 \cite{yang2025qwen3}.

\noindent\textbf{Mask Settings.} For both FFHQ and DIV2K, we adopt LaMa’s \cite{lama} strategy with polygonal thick- and thin-stroke masks, and additionally include rectangular masks covering half of the image. Masks are randomly sampled from these configurations to ensure a diverse evaluation.

\noindent\textbf{Additional Metrics.} In addition to perceptual quality metrics, we report LPIPS and SSIM to assess consistency, including results from the BrushBench evaluation. For FFHQ, we additionally report FID \cite{fid}.

\section{Analysis of the Inversion Effect}
We analyze the effect of the inversion network to explain why initializing from well-aligned noise yields more coherent and consistent outputs. Our intuition is that such noise encodes the blending trajectory and preserves background information, thereby enabling smoother blending during the denoising process.

\begin{figure}[!t]
  \centering
  \includegraphics[width=1.0\linewidth]{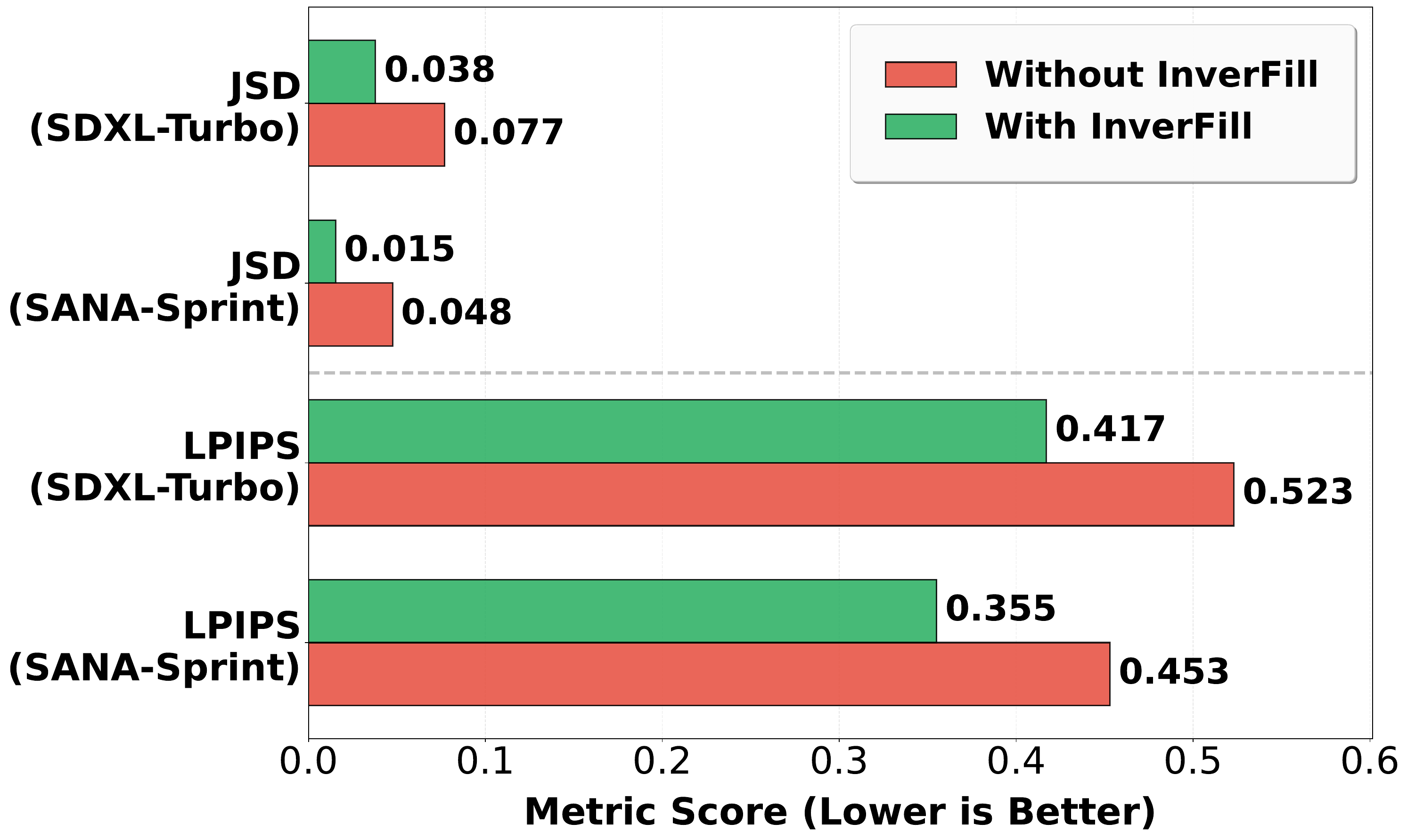}
  \caption{\textbf{Quantitative analysis of inversion effects.} We report LPIPS in background regions at intermediate timesteps and JSD with respect to the Gaussian prior. Lower values indicate better alignment. \textcolor{red}{Red} bars denote results without InverFill, while \textcolor{green}{Green} bars denote results with InverFill.}
  \label{fig:analysis}
\end{figure}

To further validate this observation, we compute LPIPS between the predicted $x_0$ in background regions at intermediate timesteps and the input image, and report the results in \cref{fig:analysis}. We observe that initialization with well-aligned noise consistently yields significantly lower LPIPS than random initialization, supporting our hypothesis. Moreover, \cref{fig:analysis} provides insight into the effectiveness of the Gaussian regularization loss: the Jensen–Shannon divergence (JSD) with respect to the Gaussian distribution is substantially reduced when this regularization is applied, leading to better-aligned latent noise while also satisfying the required Gaussian distribution for diffusion models, and consequently yielding stable and coherent reconstructions.

\section{Failure Cases}

\begin{figure}[!t]
  \centering
  \includegraphics[width=1.0\linewidth]{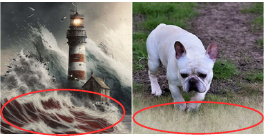}
  \caption{Representative failure cases of our method. While InverFill improves overall coherence, it may produce color inconsistencies between the inpainted regions and the background.} 
  \label{fig:Failure_cases}
  \vspace{-5mm}
\end{figure}

We report representative failure cases in \cref{fig:Failure_cases}. Overall, the main limitation of our method stems from color inconsistencies between the inpainted region and the background.

\section{More Qualitative Results}
\label{sec:morequal}
\begin{figure*}[t]
  \centering
  \includegraphics[width=\linewidth]{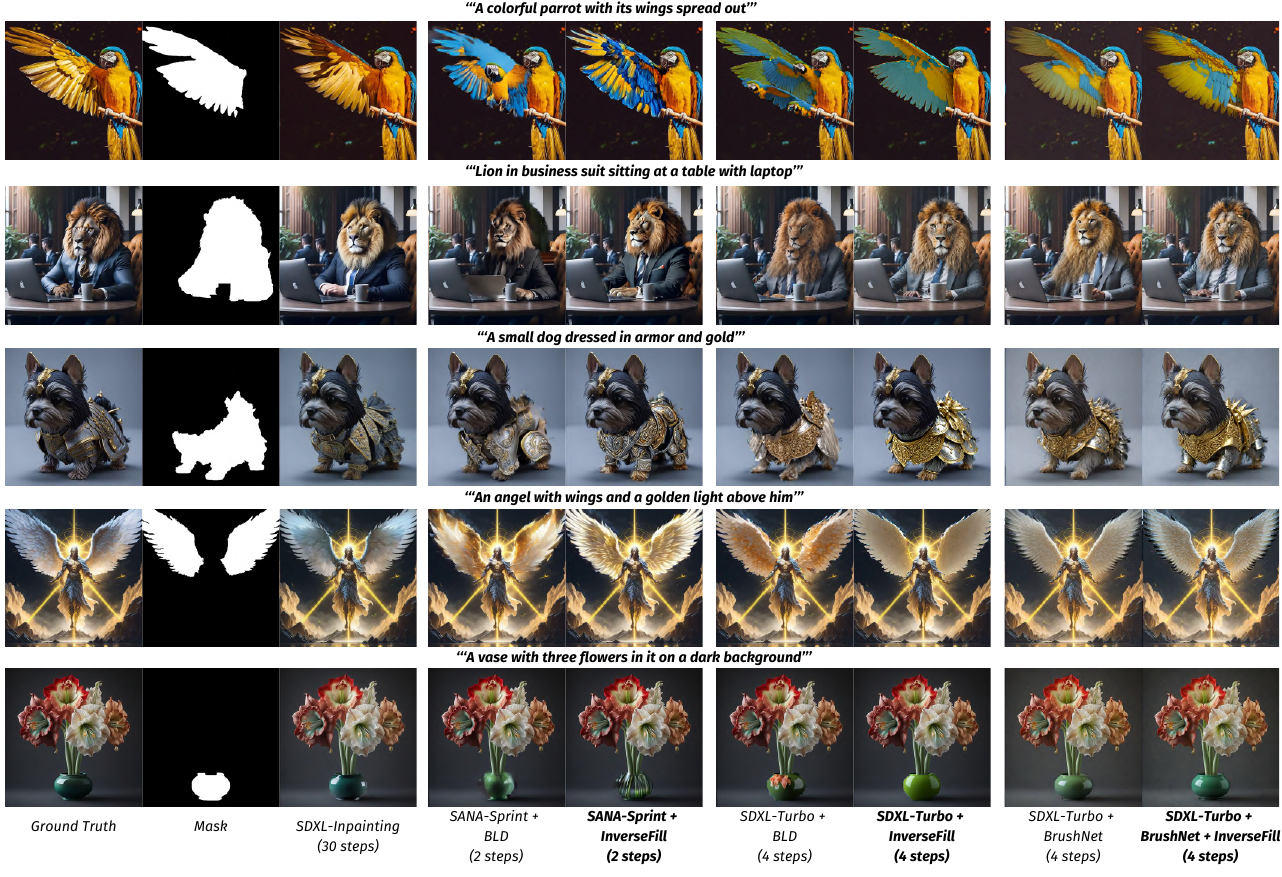}
  \caption{More qualitative comparison on BrushBench \textit{(Zoom in for best view)}} 
  \label{fig:More_Qual_Brushbench}
\end{figure*}

\begin{figure*}[t]
  \centering
  \includegraphics[width=\linewidth]{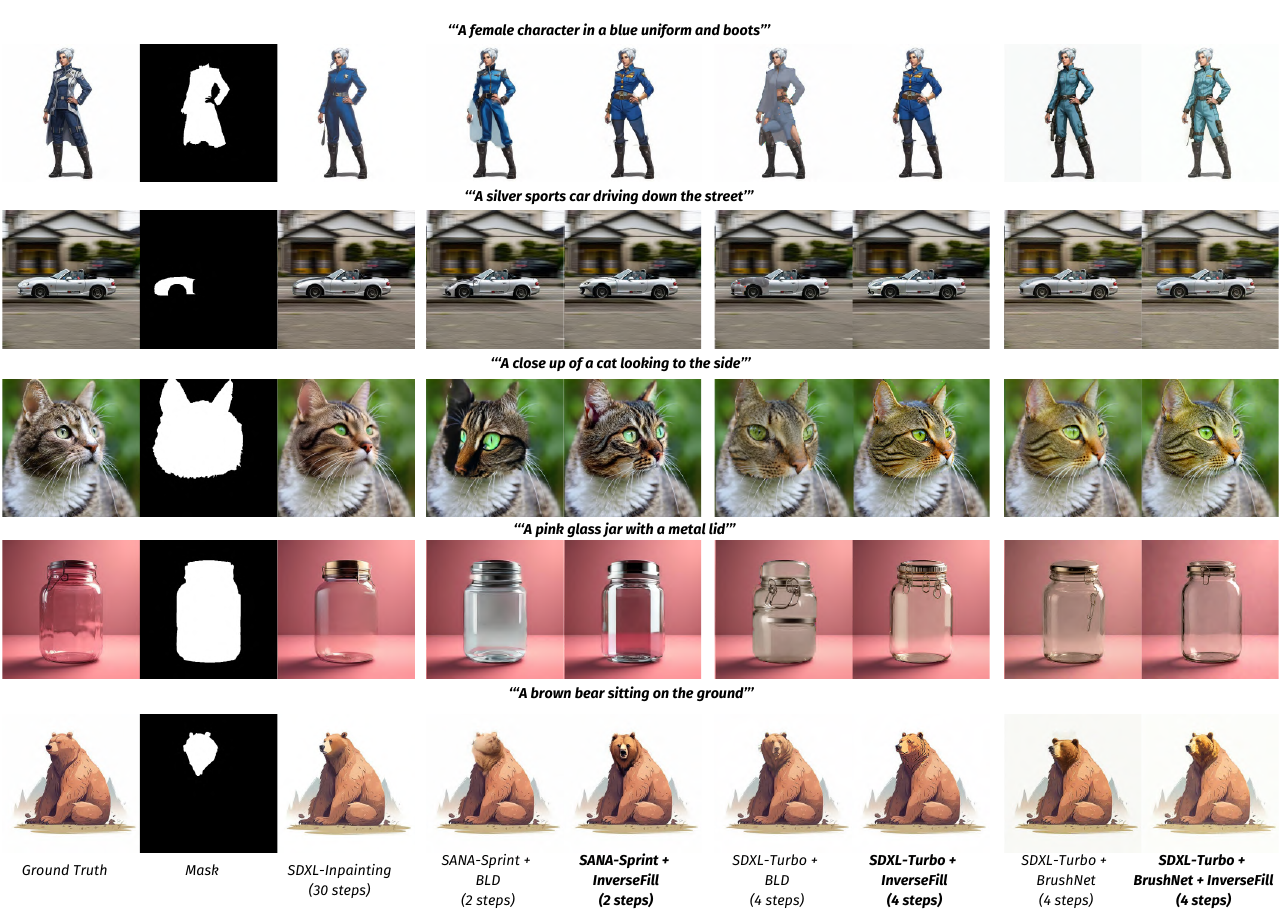}
  \caption{More qualitative comparison on BrushBench \textit{(Zoom in for best view)}} 
  \label{fig:More_Qual_Magicbench_2}
\end{figure*}

To provide a comprehensive visual comparison of InverFill, \cref{fig:More_Qual_Brushbench,fig:More_Qual_Magicbench_2} present an expanded set of qualitative results, further illustrating the improvements in coherence and background harmonization highlighted in our work.

\end{document}